\documentclass[journal]{IEEEtai}

\usepackage[colorlinks,urlcolor=blue,linkcolor=blue,citecolor=blue]{hyperref}

\usepackage{color,array}

\usepackage{graphicx}
\usepackage{amsmath}
\usepackage{hyperref}

\usepackage{url}
\usepackage{multirow}
\usepackage{amssymb}
\usepackage[table,xcdraw]{xcolor}
\usepackage{ragged2e}

\begin{document}

\title{AI Learning Algorithms: Deep Learning, Hybrid Models, and Large-Scale Model Integration} 

\author{
    Noorbakhsh Amiri Golilarz, 
    Elias Hossain, 
    Abdoljalil Addeh, 
    and Keyan Alexander Rahimi
    \thanks{Noorbakhsh Amiri Golilarz is with the Department of Computer Science, The University of Alabama, Tuscaloosa, AL 35487, USA (e-mail: noor.amiri@ua.edu).} \thanks{Elias Hossain is with the Department of Computer Science and Engineering, Mississippi State University, MS 39762, USA (e-mail: mh3511@msstate.edu).}
    \thanks{Abdoljalil Addeh is with the Department of Biomedical Engineering, Schulich School of Engineering, University of Calgary, AB, Canada (e-mail: abdoljalil.addeh@ucalgary.ca).}
    \thanks{Keyan Alexander Rahimi is with the Department of Computer Science, Brown University, Providence, RI 02912, USA (e-mail: keyan.rahimi@brown.edu).}
    \thanks{Corresponding author: Elias Hossain (e-mail: mh3511@msstate.edu).}
}





\maketitle

\begin{abstract}
In this paper, we discuss learning algorithms and their importance across various applications, emphasizing their ability to identify key patterns and features in a straightforward, easy-to-understand manner. We review the main concepts of Artificial Intelligence (AI), Machine Learning (ML), Deep Learning (DL), and hybrid models. Key subsets of ML algorithms, such as supervised, unsupervised, and reinforcement learning, are also covered. These techniques are essential for tasks such as prediction, classification, and segmentation. Convolutional Neural Networks (CNNs), widely used in image and video processing, are explored in detail, including their architecture and how they can be integrated with ML algorithms to build hybrid models. This paper also explores the vulnerability of learning algorithms to adversarial noise, which can lead to misclassification. Furthermore, we discuss the integration of learning algorithms with Large Language Models (LLMs) to generate coherent responses in domains such as healthcare, marketing, and finance by learning and extracting meaningful patterns from large datasets. Finally, we discuss the next generation of learning algorithms, emphasizing the potential for a unified, adaptive, and dynamic network capable of performing a wide range of tasks. Overall, this article provides a concise overview of learning algorithms, highlighting their current capabilities, applications, and future directions.

\end{abstract}



\section{Introduction}
Learning algorithms are computational techniques that autonomously refine their performance by accumulating experience, detecting patterns within data, and making informed predictions or decisions based on that data. These algorithms are integral components of Artificial Intelligence (AI), Machine Learning (ML) and Deep Learning (DL) frameworks.

\begin{figure}[ht]
\centering
\includegraphics[width=0.95\linewidth]{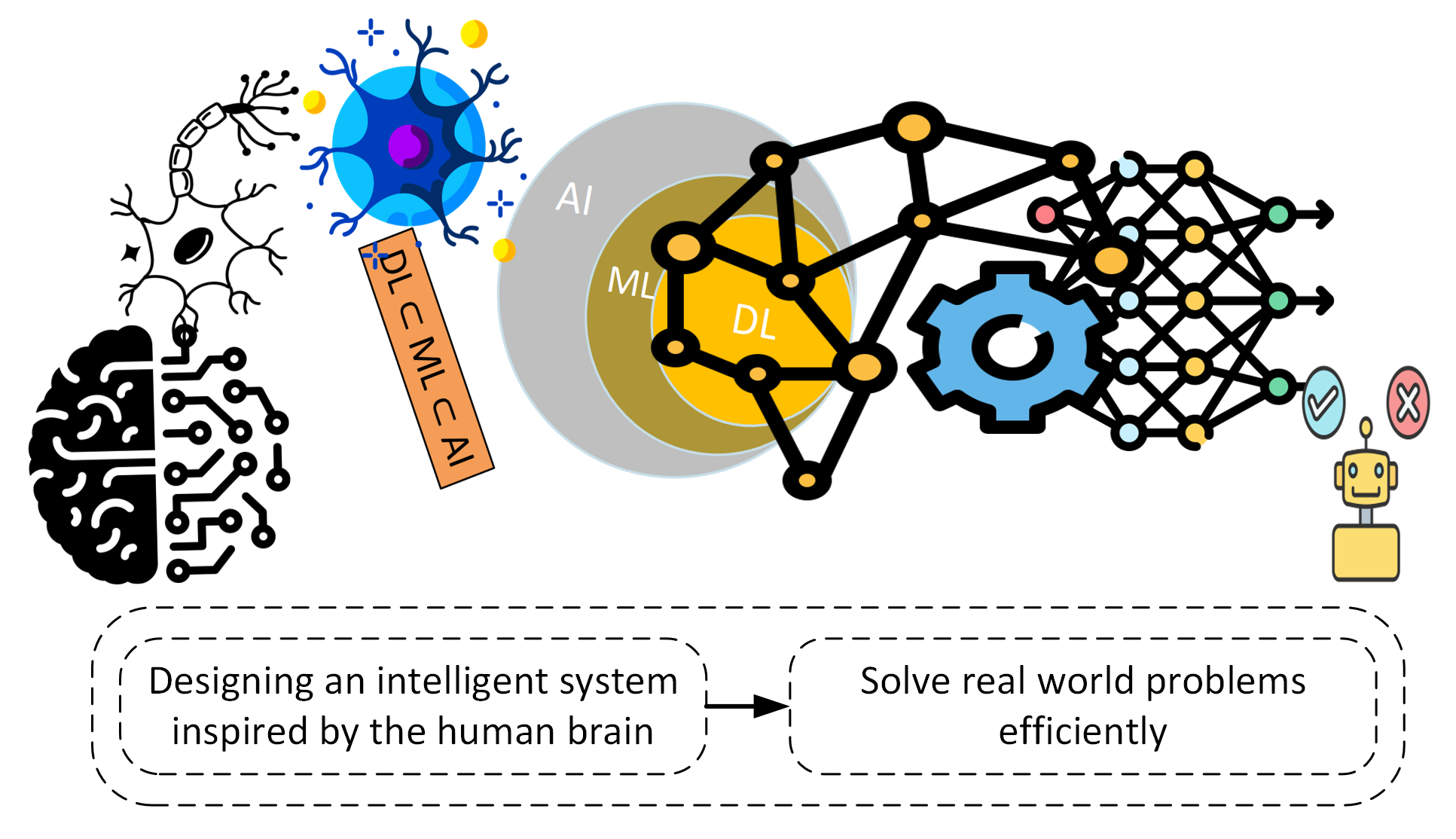}
\caption{AI system and the relationship between AI, ML and DL.}
\label{fig:intro-fig}
\end{figure}

Artificial Intelligence (AI) refers to the use of machines to perform tasks traditionally requiring human intelligence, such as decision-making or object recognition (see Fig. \ref{fig:intro-fig}). In recent years, AI technologies have rapidly evolved, becoming deeply embedded in everyday life. This growth has been fueled by significant advancements in science and technology \cite{liu2021we}, positioning AI and its multidisciplinary applications among the most influential subjects of the future. AI is now ubiquitous across major technology-driven platforms such as Google, YouTube, and Amazon. It has found widespread use in domains including disease diagnosis, early tumor prediction, recommendation systems, image and signal processing, and computer vision. Unlike traditional approaches, AI empowers experts to solve complex problems more efficiently by leveraging extensive computer programming that enables machines to emulate human behavior \cite{harika2022review}. This represents a full integration between human cognition and computer-based technologies \cite{liu2021we}.

Prediction and classification are among the most essential cognitive tasks that AI-based systems aim to replicate. Achieving high accuracy in these tasks requires sophisticated algorithms capable of mimicking human skills and decision-making abilities \cite{sumari2021prediction}. As illustrated in Fig. \ref{fig:intro-fig} \cite{chollet2021deep}, Machine Learning (ML) is a sub-field of AI, while Deep Learning (DL) is a further subset within ML. These learning paradigms form the computational backbone of modern AI systems. Brain-inspired AI approaches, particularly Deep Learning and Cognitive Neuroscience, further refine AI’s ability to process information similarly to the human brain (see Fig. \ref{fig:intro-fig}). These approaches leverage neural network architectures that emulate biological neurons, enhancing AI’s ability to tackle complex, real-world problems efficiently. By adopting strategies inspired by the brain’s structure and functionality, these methods enable AI systems to exhibit adaptive learning, robust pattern recognition, and enhanced problem-solving capabilities, making them a crucial advancement in intelligent system design.

ML involves developing statistical models that can learn from data and generalize to unseen data, enabling systems to execute tasks without explicit instructions \cite{wiki:Machine_Learning}. It plays a crucial role in fields like computer vision, natural language processing, and robotics. The concept of training data—data used to build a predictive model—is foundational to ML systems. ML applications span from speech and image recognition to personalized recommendation systems and predictive analytics in healthcare.

DL, a specialized branch of ML, employs neural networks to perform tasks by automatically learning hierarchical feature representations from large datasets. Inspired by the structure of biological neurons, DL models process inputs such as text, images, or audio through deep architectures. The accuracy and effectiveness of DL models are influenced by factors such as dataset size, network architecture, optimization techniques, and computational resources. DL has been instrumental in advancing technologies like natural language processing, autonomous driving, object detection, and virtual assistants.

Despite their success, learning algorithms still face challenges. In critical domains like healthcare, high-quality annotated datasets are essential. Moreover, the interpretability and complexity of these models can hinder their adoption. As DL models often require substantial computational resources, future developments may focus on lightweight algorithms and efficient hardware solutions to improve scalability and accessibility.

Although numerous review articles have explored learning algorithms \cite{bottou1992local} \cite{japkowicz2011evaluating} \cite{lakshmivarahan2012learning}, most focus exclusively on either classical machine learning techniques or deep learning models. In contrast, this review provides an integrative and up-to-date analysis—ranging from traditional ML and DL to emerging paradigms such as Explainable AI, adversarial robustness, federated learning, and Large Language Models. This comprehensive scope enables readers to understand both foundational concepts and cutting-edge trends within a single framework, making this review relevant to both academic and applied research communities.

This article aims to provide a systematic and unified overview of the wide range of learning algorithms available in the AI domain. As technology advances rapidly and domain-specific expertise becomes more widespread, researchers often face fragmented knowledge. This review integrates various learning paradigms to serve as a comprehensive resource for both early-stage learners and experienced researchers. 
It also illustrates how learning algorithms change and adapt in practical applications by discussing both established and new directions.


The key contributions of this article have been outlined as follows:

\begin{itemize}
    \item A comprehensive overview of both classical and modern learning algorithms is provided in a straightforward manner, with examples that help unify the scattered knowledge in the field.
    
    \item A systematic comparison between traditional machine learning paradigms and contemporary approaches, including deep learning and hybrid models, has been conducted.
    
    \item Emerging topics such as Explainable AI (XAI), adversarial robustness, federated learning, and Large Language Models (LLMs) have been incorporated to reflect current advancements.
    
    \item Fragmented knowledge across different domains has been bridged by integrating various learning paradigms into a cohesive review.
    
    \item Practical applications in sensitive domains, such as healthcare  and natural language understanding, have been emphasized to highlight real-world impact.
\end{itemize}

The rest of the paper is organized as follows. In Section \ref{sec:machine-learning}, we discuss various machine learning algorithms, including supervised learning, unsupervised learning, reinforcement learning, semi-supervised learning, federated learning, feature engineering/learning, transfer learning, and ensemble learning. In Section \ref{sec:deep-learning}, we explain deep learning algorithms such as Convolutional Neural Networks (CNNs), Recurrent Neural Networks (RNNs), Long Short-Term Memory networks (LSTMs), Generative Adversarial Networks (GANs), the transformer architecture, and backpropagation. In Section \ref{sec:hybrid-models}, we demonstrate hybrid models. In Section \ref{sec:xai}, we provide an explanation of the explainable artificial intelligence methods. In Section \ref{sec:fooling-learning}, we discuss fooling the learning algorithms into misclassification. In Section \ref{sec:llm-integration}, we show how LLMs are integrated into various sectors. In Section \ref{sec:real-world-applications}, we depict real-world applications, and in Section \ref{sec:future-of-learning}, we explore the future of learning algorithms. Finally, in Section \ref{sec:conclusion}, we wrap up the manuscript.

\section{Machine Learning}
\label{sec:machine-learning}
Machine learning is categorized as a subset of artificial intelligence that encompasses probability theory, statistics, approximation theory, convex analysis, and algorithm complexity which deals with learning from the data to act like human beings, and to analyze and process the given tasks efficiently. It has a wide variety of applications in biomedical, engineering, science, agriculture, animal sciences, neuroscience, remote sensing, etc. Those equipped with ML can run decision-making and performance analysis much faster than they ever could have before.  However, this section is dedicated to some fundamental computer vision tasks like segmentation, classification, and detection, as they are highly applicable in numerous practical scenarios.

\begin{figure}[ht]
\centering
\includegraphics[width=0.8\linewidth]{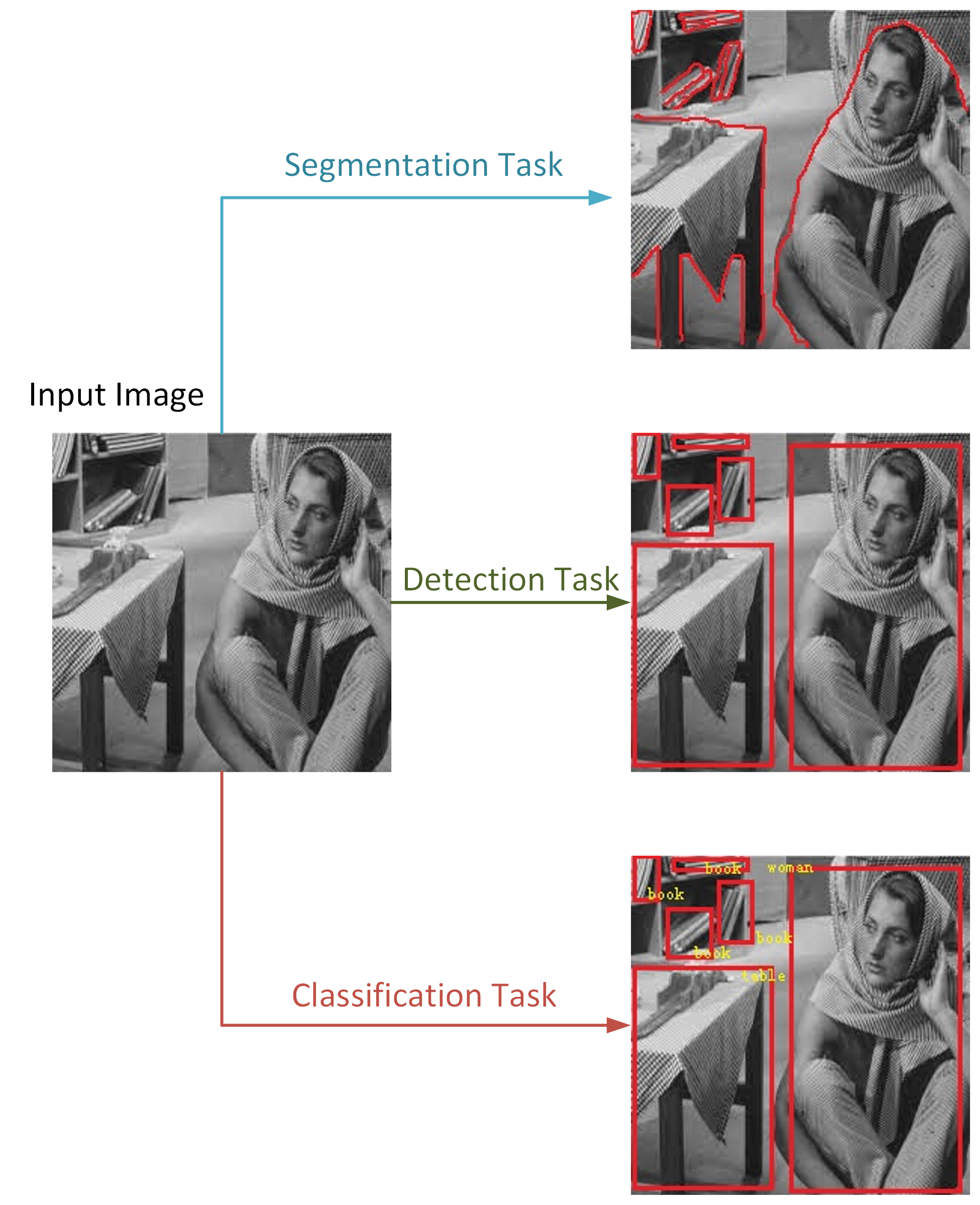}
\caption{Some important machine learning tasks. The main purpose of learning-based algorithms is extracting higher-level information from input data. More progress in ML approaches much better image classification, object detection, and segmentation.}
\label{fig:image-seg}
\end{figure}

Segmentation is one of the critical steps to be done prior to further image analysis and processing.
The segmented features are supposed to be fed to the algorithm for recognition and classification. Some published research has shown that ML can process these tasks properly \cite{mahadevkar2022review}. Fig. \ref{fig:image-seg} shows some of these tasks for Barbara test image \cite{chang2000adaptive}. In this section, we also discuss some important concepts and techniques in ML.

\subsection{Supervised Learning}

Supervised learning is a subset of ML in which the training can be processed on the labeled dataset. In this case, the algorithm can learn from the labeled input data. If it is provided with a new set of examples, then based on the learned features from the training stage, the machine can predict the outcome accurately. This characteristic allows supervised learning to solve various problems with high accuracy. The two sub-categories of supervised learning are classification (SVM and neural nets) and regression (logistic and linear regression).

\begin{figure}[ht]
\centering
\includegraphics[width=0.8\linewidth]{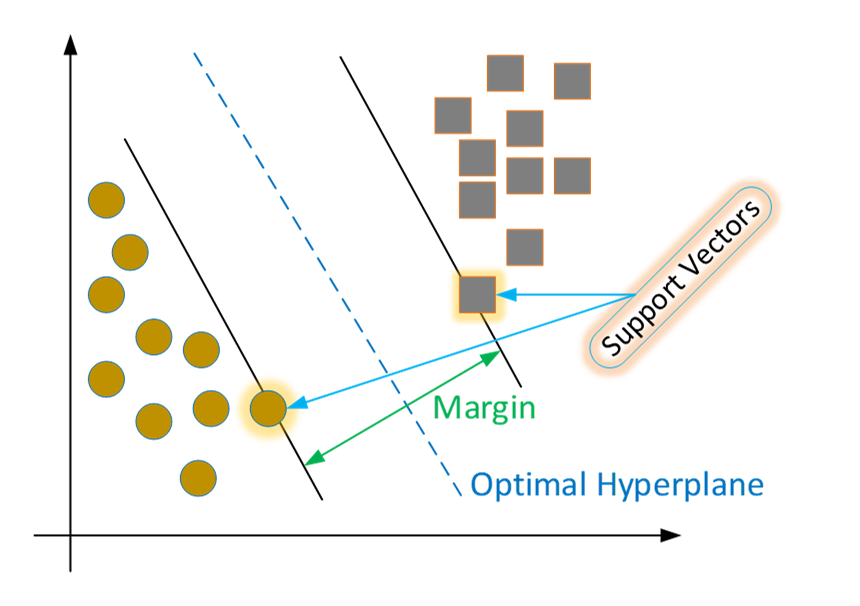}
\caption{Illustrating the SVM model, showcasing the support vectors and optimal hyperplane that separate different classes in the feature space \cite{cortes1995support}.}
\label{fig:svm_model}
\end{figure}

As mentioned above, SVM is a supervised classification algorithm. According to Fig. \ref{fig:svm_model}, let’s consider a training dataset of \( n \) points as \( (x_1, y_1), \ldots, (x_n, y_n) \) and also consider \( w^T x + b = 0 \) as the hyperplane separating two classes linearly \cite{cortes1995support}. We can write the optimization problem as:


\begin{align}
\min_{w, b} \frac{1}{2} w^T w \quad & \text{subject to} \nonumber \\
y_i (w^T x_i + b) & \geq 1, \label{eq:svm}
\end{align}
where \( y_i \in \{-1, 1\} \) is the class label, \( w \) is the weight, and \( b \) is the bias.

On the other hand, let’s imagine the data points cannot be separated linearly or without misclassification. In this case, we will add one more constraint \cite{cortes1995support}. We are dealing with minimizing the misclassification error. The new optimization problem can be written as:


\begin{align}
\min \frac{1}{2} \|w\|^2 + C \sum_{i=1}^n \phi_i \quad & \text{subject to} \nonumber \\
y_i (w^T x_i + b) & \geq 1 - \phi_i, \quad \phi_i \geq 0, \quad i = 1, \ldots, n, \label{eq:support_vector_machine}
\end{align}
where \( C \) is the regularization parameter and \( \phi_i \) is the slack variable.

\begin{figure}[ht]
\centering
\includegraphics[width=0.6\linewidth]{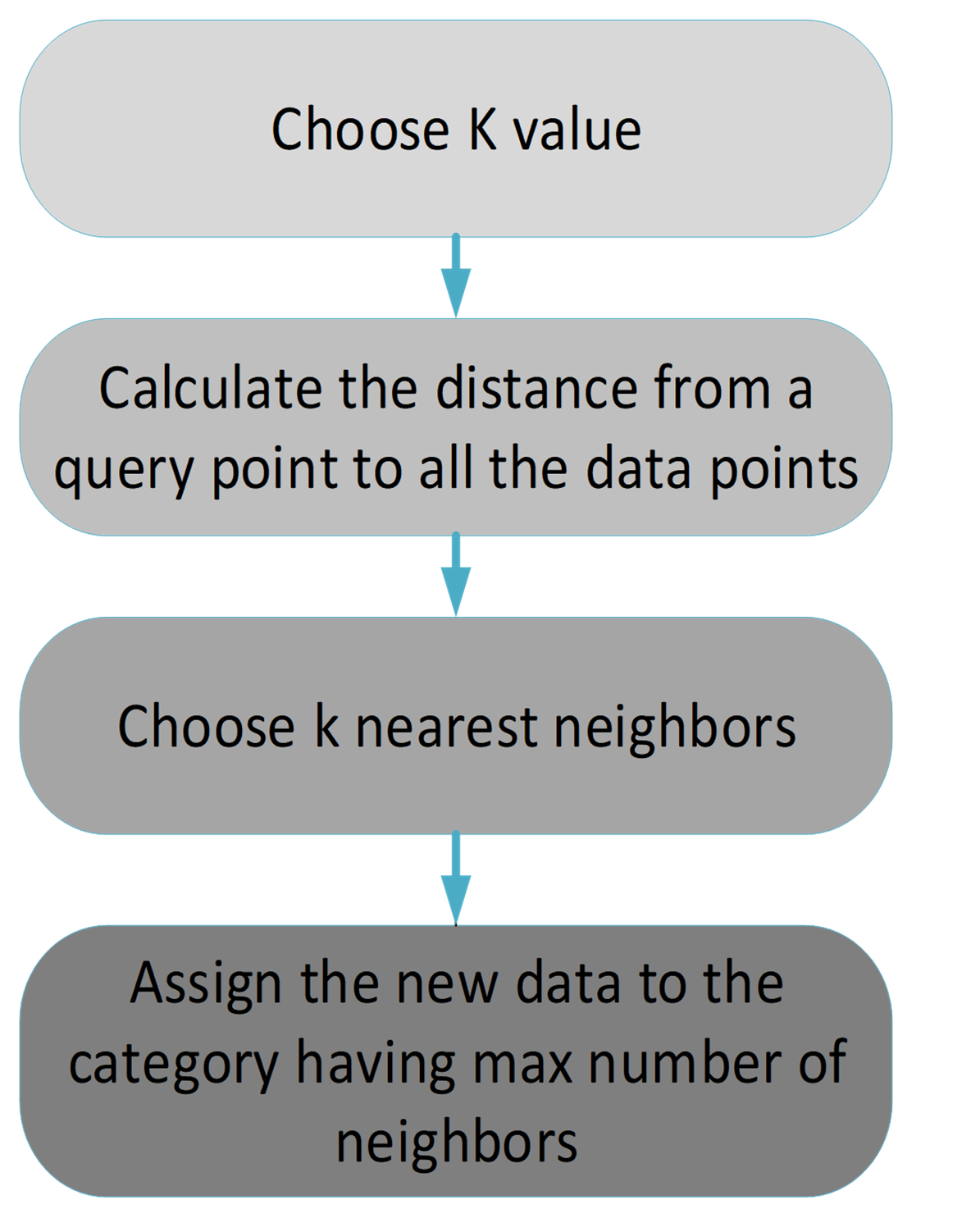}
\caption{Overview of the KNN algorithm, outlining the steps involved in categorizing an unclassified data point based on majority voting from its nearest neighbors \cite{uddin2022comparative}.}
\label{fig:knn-main-step}
\end{figure}

K-NN is categorized as a supervised algorithm that can be utilized for both classification and regression. The overall steps of KNN is illustrated in Fig. \ref{fig:knn-main-step}. Consider several categories (sets of classified points \cite{cover1967nearest}) and the number of $k$ neighbors (three) for an unclassified data point $q_1$. Assume one neighbor belongs to class $O$, while the other two belong to class $X$. In this circumstance, $q_1$ will be classified as class $X$ because the majority of our neighbors are members of that class. This classification can be accomplished with simple majority vote or distance-weighted voting \cite{Cunningham_2021}.

\subsection{Unsupervised Learning}

Unsupervised learning deals with input data that are not labeled at all. The input is unsorted information or data and we do not have any idea, clue or guidance about the input characteristics and features. In this case, grouping can be processed based on the shape, differences, similarities and different patterns of the data. In terms of complexity, it is a computationally complex model. Unlike supervised learning, the number of classes is unknown. Unsupervised learning can be used for clustering, feature learning, anomaly detection, and dimensionality reduction \cite{mahadevkar2022review}. 
Clustering is one of the important types of un-supervised machine learning algorithms which group and categorize unlabeled data. It is similar to classification, but the only difference is in the dataset. In classification we are dealing with labeled data but in clustering we are working with un-labeled dataset. To provide a clearer understanding of this concept, the general framework of clustering is shown in Fig. \ref{fig:super-semi-rein} (b).



\begin{figure}[ht]
\centering
\includegraphics[width=0.7\linewidth]{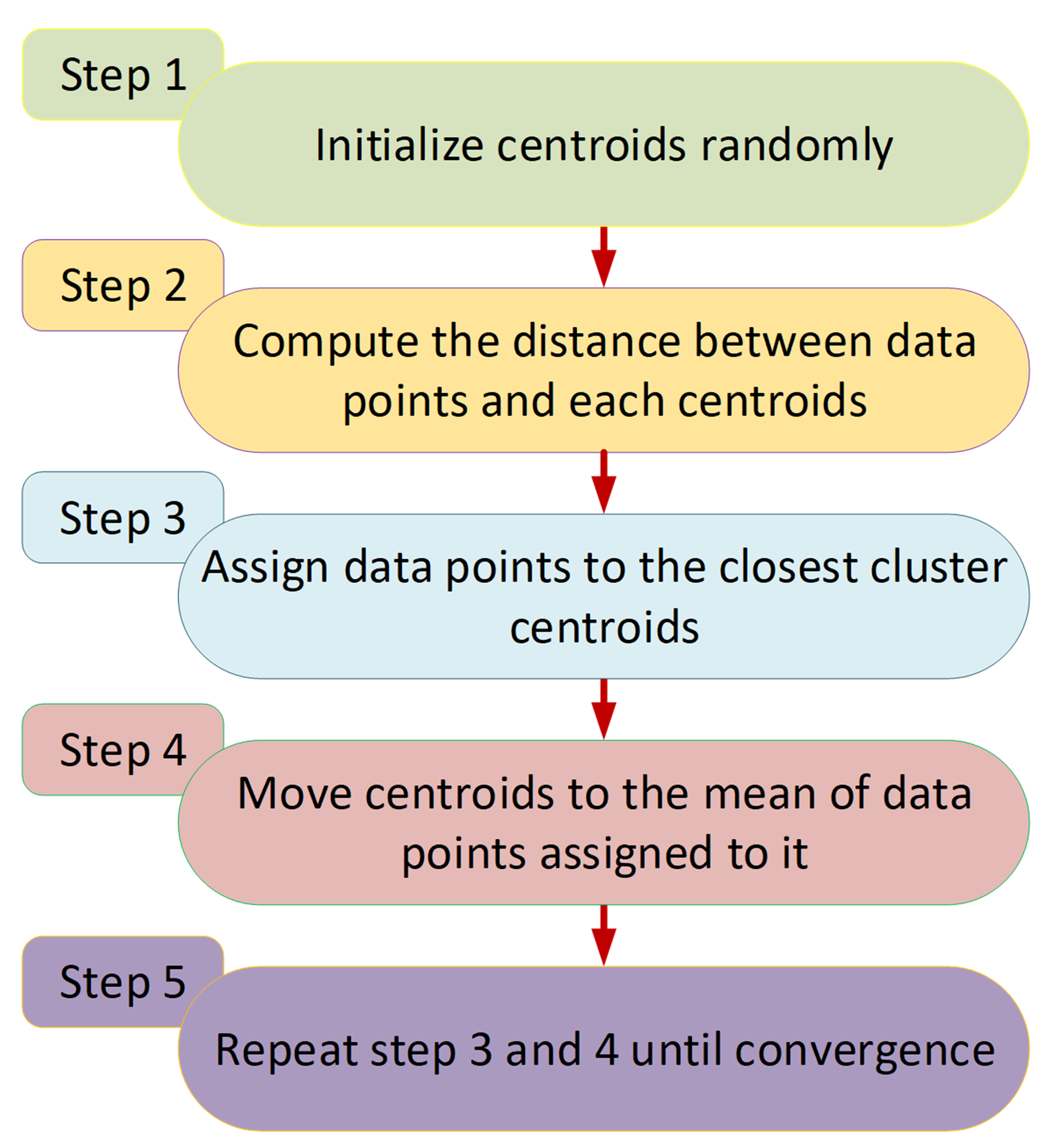}
\caption{Key steps of the K-Means clustering algorithm, illustrating the process of grouping data points into distinct clusters based on their features.}
\label{fig:main-steps-kmean-clustering}
\end{figure}

K-means clustering is a type of unsupervised learning algorithm which is working based on randomly initializing the centroids, computing the distances among given data points and centroids, and assigning the sample points to the new clusters. The main steps of the k-means clustering are depicted in Fig. \ref{fig:main-steps-kmean-clustering} \cite{piech_kmeans}. 

\begin{figure}[ht]
\centering
\includegraphics[width=0.9\linewidth]{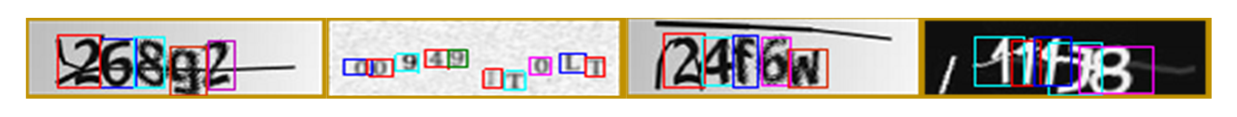}
\caption{ Samples of segmentations based on k-means clustering}
\label{fig:k-means-seg}
\end{figure}

In Fig. \ref{fig:k-means-seg}, we try to find the best place to separate the image after filtering with k-means clustering method. The images have been collected from Kaggle \cite{wilhelmy_rosas_captcha}. In this method as briefly described in \cite{zheng2011character}, we can use the final image in the form of a matrix. We select several random starting points in the matrix. Using the k-means method, we divide the points inside the image into several clusters \cite{zheng2011character}. The center point of each cluster can also be considered as the center of each character. 

\subsection{Reinforcement Learning}
\begin{figure}[ht]
\centering
\includegraphics[width=0.9\linewidth]{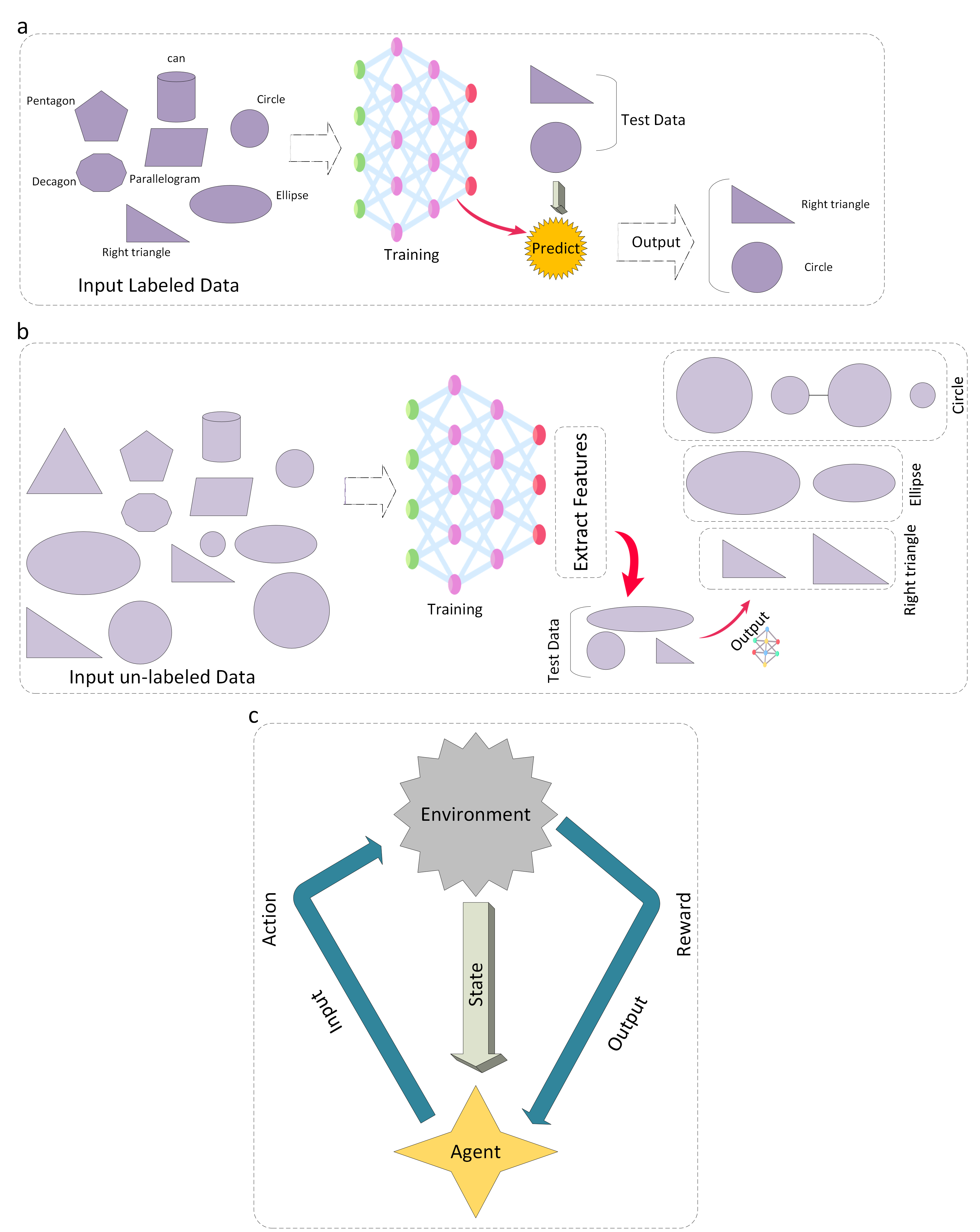}
\caption{Illustrative frameworks of different machine learning paradigms: (a) supervised learning, which uses labeled data for training; (b) unsupervised learning, which aims to find patterns from unlabeled data; and (c) reinforcement learning \cite{mahadevkar2022review}, where an agent interacts with an environment to maximize cumulative rewards.}
\label{fig:super-semi-rein}
\end{figure}

Reinforcement learning (RL) is another subset of machine learning in which the agents are allowed to learn from their own experiences and errors in an environment leading to maximized total cumulative reward and making a balance between exploration and exploitation which are the main goals of RL \cite{wikipedia_reinforcement_learning}. As it is mentioned above, the main aim of unsupervised learning is to deal with unsorted data, differences, and similarities among them. Also, the main target in supervised learning is dealing with labeled data for better prediction and classification. The framework of supervised, unsupervised, and reinforcement learning are depicted in Fig. \ref{fig:super-semi-rein}.

\subsection{Semi-supervised Learning}
\
Some drawbacks in supervised and unsupervised learning algorithms like manually labeling the data and limited spectrum of applications, respectively \cite{javatpoint_semi_supervised_learning} are the main motivations of developing semi-supervised learning. 

\begin{figure}[ht]
\centering
\includegraphics[width=0.55\linewidth]{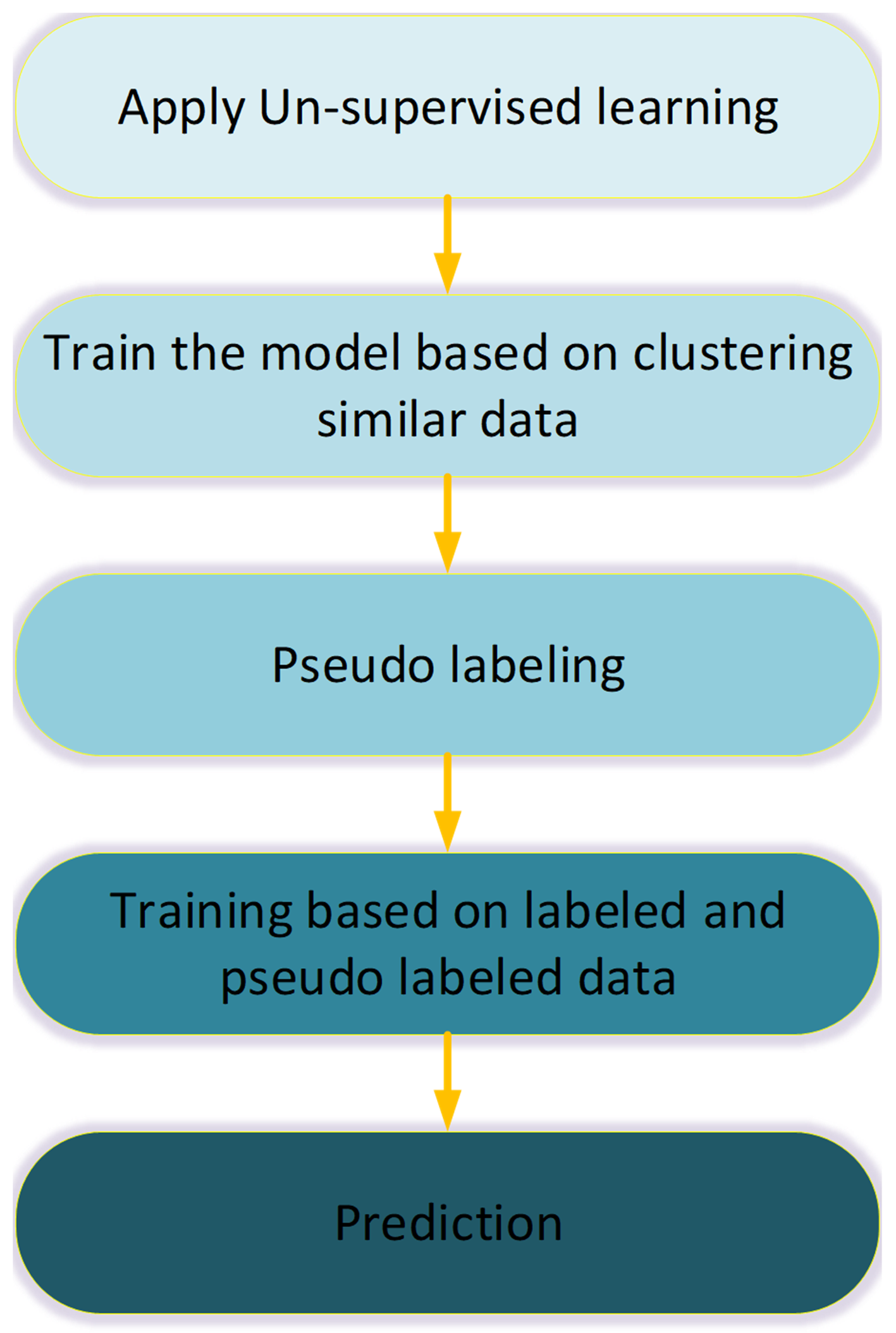}
\caption{ Overview of the main steps in semi-supervised learning \cite{mahadevkar2022review}.}
\label{fig:semi-supervised-step}
\end{figure}

In this type of learning, we are dealing with both labeled and unlabeled data. As seen in Fig. \ref{fig:semi-supervised-step}, the initial step here is clustering based on the similarity in the input data. It means that unsupervised learning is used first to cluster similar data and use it to train the model. Then this information can be used to label the remaining unlabeled data \cite{mahadevkar2022review}. This process is called pseudo-labeling. Next, train the model based on these combined labeled and pseudo-labeled data \cite{mahadevkar2022review}\cite{javatpoint_semi_supervised_learning} which results in prediction with better accuracy.

\subsection{Federated Learning}
\label{federate-learning}

Federated Learning (FL) offers a revolutionary training strategy for creating individualized models while preserving user privacy \cite{zhang2021survey} \cite{vikash2024federated}. With the introduction of artificial intelligence chipsets, client devices' processing resources have increased, gradually shifting model training from a central server to terminal devices. Additionally, this approach provides a privacy protection mechanism that leverages the processing capabilities of terminal devices to train models, preventing private information from being leaked during data transmission. It also fully utilizes the vast dataset resources available from smartphones and other devices \cite{zhang2021survey}.

\begin{figure}[h]
\centering
\includegraphics[width=0.95\linewidth]{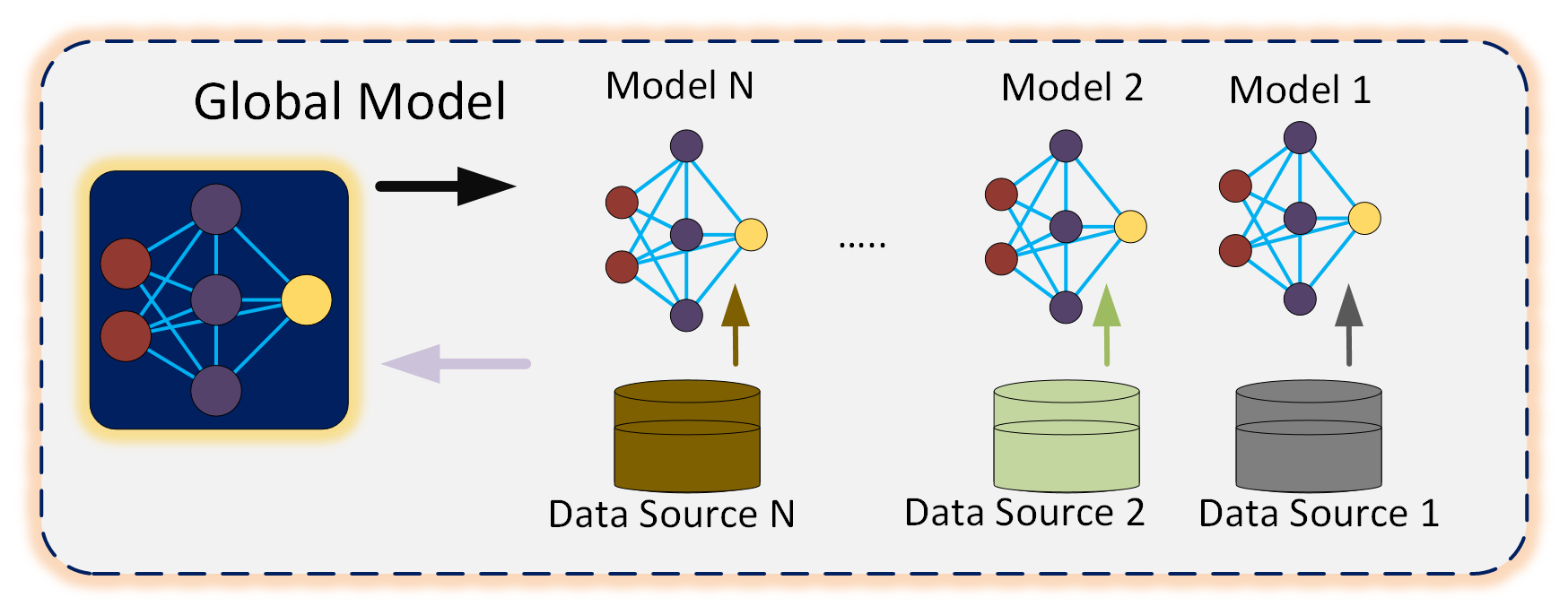}
\caption{The main working procedure of federated learning.}
\label{fig:federate-learning}
\end{figure}

Notably, we demonstrated how federate learning operates in Fig. \ref{fig:federate-learning}, where several models are trained on various data sources while preserving security \cite{kumar2021blockchain}. It is demonstrated that each model is trained locally on the relevant data source, and after that, it sends its updated parameters to a central server, which aggregates them to create a global model. After that, the individual models receive further training from this global model, which enhances their functionality and ability to generalize. This collaborative learning is made possible through this iterative technique without disclosing private information.

Additionally, a major function of this training paradigm is to ensure user privacy, distinguishing it significantly from typical privacy protection techniques used in big data, such as differential privacy and k-anonymity; moreover, federated learning primarily safeguards privacy by transmitting encrypted processing parameters, preventing attackers from obtaining source data. These measures ensure data-level privacy and compliance with the General Data Protection Regulation (GDPR) and other regulations \cite{zhang2021survey}. 

Depending on data distribution, FL can be classified as horizontal, vertical, or federated transfer learning. Horizontal federated learning is appropriate when user features of two datasets overlap significantly but the users overlap minimally \cite{zhang2021survey}. Vertical federated learning is applicable when user features overlap slightly, but users overlap significantly. When users and features rarely overlap, transfer learning compensates for the lack of data or tags \cite{zhang2021survey}. Moreover, this technique is comparable to multiparty computing and distributed machine learning, which includes distributed model outcomes, distributed data storage, and distributed computing activities \cite{zhang2021survey}. In distributed machine learning, the parameter server helps increase training speed by distributing data across working nodes and managing resources via a trusted central server. Furthermore, it allows each worker node to retain its own data while participating in model training.

On the other hand, a critical aspect of this paradigm in ensuring privacy is allowing users complete control over their local data, emphasizing data owners' privacy \cite{zhang2021survey}. There are two types of privacy protection measures used in a federated environment: encryption algorithms such as homomorphic encryption, and secure aggregation techniques \cite{zhang2021survey}. Another common approach is incorporating the noise of differential privacy into model parameters. For instance, Google proposes a method using a combination of secure convergence and differential privacy \cite{bonawitz2017practical}. Other studies have utilized solely homomorphic encryption to ensure privacy \cite{liu2020secure}.

\subsection{Feature Learning}

Feature learning is an approach that allows machines to learn the characteristics and representations required for feature selection, recognition, and classification \cite{bengio2013representation, wikipedia_feature_learning}, resulting in much better representations, making feature extraction easier and more accurate \cite{bengio2013representation}. In contrast, manually extracting features may not yield an optimal feature set or accurate predictions \cite{mahadevkar2022review}. Consequently, the following methods serve as a replacement for manual feature extraction: autoencoders, principal component analysis (PCA), bag of words (BoW), term frequency-inverse document frequency (TF-IDF), and image processing techniques.

Firstly, the autoencoder concept is based on learning from the coding of the original data sets to create new and more powerful features. It accomplishes this by training a neural network to replicate its input, forcing it to identify and exploit data structures \cite{domino_feature_extraction}. It minimizes dimensionality and extracts key features from data, resulting in more effective machine-learning models. Secondly, large data sets' dimensionality is decreased while maintaining as much information as possible using the PCA-based feature extraction technique. Principal Component Analysis highlights variation and identifies significant trends and connections among the dataset's variables \cite{domino_feature_extraction}. Next, BoW is a useful method in Natural Language Processing (NLP) that allows words (also known as features) to be retrieved from a text and categorized based on how often they are used. Every document is represented by a vector of word counts, and the word count is sent into machine learning algorithms as input \cite{domino_feature_extraction}. 

\begin{figure}[h]
\centering
\includegraphics[width=\linewidth]{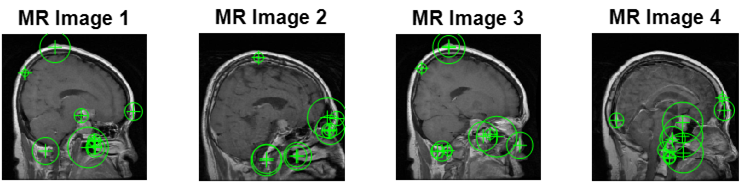}
\caption{Demonstrating the detection of key points using SURF (Speeded-Up Robust Features) technique.}
\label{fig:surf-keypoints-images}
\end{figure}

Then, TF-IDF, an NLP-based feature extraction technique, is an extension of BoW that uses a numerical statistic to represent a word's importance to a document inside a collection or corpus \cite{domino_feature_extraction}. In contrast to BoW, it takes into account all of the other texts in the corpus in addition to a word's frequency in a single document. This makes up for the fact that some terms are used more frequently than others overall. Finally, to recognize and isolate important features or patterns in a picture, image processing algorithms analyze raw data \cite{domino_feature_extraction}. The process could be extracting properties such as color, texture, and shape, or it could be recognizing corners and edges. Then, tasks like object detection, image segmentation, and image classification can be performed using these features \cite{domino_feature_extraction}. For example, as can be seen in Fig. \ref{fig:surf-keypoints-images}, detection of key points using SURF (Speeded-Up Robust Features) technique can provide us with important features in images. The dataset is available in \cite{yeafi_brain_mri_images_2022}.


\begin{figure*}[ht]
\centering
\includegraphics[width=0.85\linewidth]{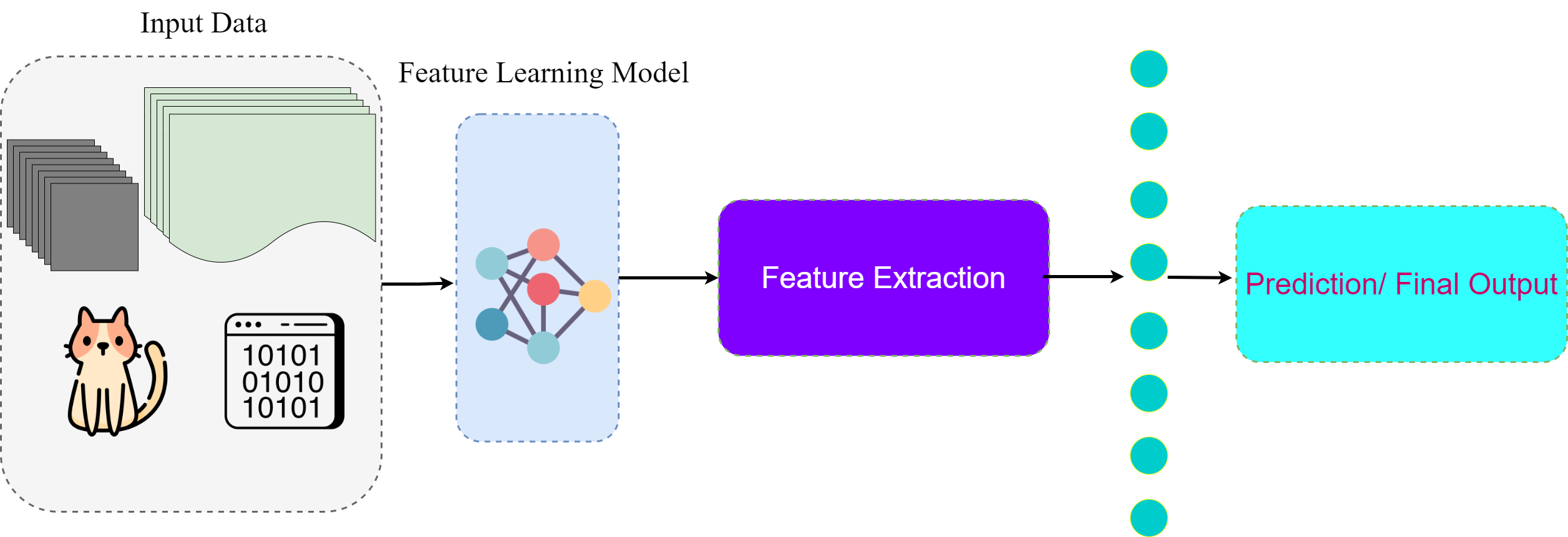}
\caption{Illustrating the feature learning in ML, allowing the models to extract and identify important features, and employ them to perform classification and prediction.}
\label{fig:feature-learning}
\end{figure*}

We illustrated the feature learning paradigm in machine learning in Fig. \ref{fig:feature-learning}. As it is shown in this figure, an initial (often sparse) feature set or raw data are input to learn implicit feature representations using a variety of techniques \cite{wiki:feature_learning}. As a result, the feature representation is richer and frequently lower dimensional, which might improve performance when applied as the input to more specialized learning tasks \cite{wiki:feature_learning}.

\subsection{Transfer Learning}

In machine learning, transfer learning is a method where knowledge acquired from one task or dataset is applied to enhance model performance on a different but related task or dataset \cite{ibm:transfer_learning}. Stated differently, this method improves generalization in a different context by applying knowledge gained in a previous context \cite{ibm:transfer_learning}. It can be used for many tasks, including deep learning model training and regression problem-solving. For situations involving deep neural networks, which need a lot of data to function well, this method is particularly appealing \cite{ibm:transfer_learning}.

Conventional learning procedures use the available labeled data to create a new model for every task. This is because, even when trying to complete a task that is similar to the first model, users must retrain a newer model from the beginning if the data distribution changes or the trained model is utilized for a completely new dataset \cite{ibm:transfer_learning}. This happens because traditional machine learning algorithms assume that training and test data originate from the same feature space. On the other hand, this method starts with models or networks that have previously been trained. To handle a new but related target task or dataset, like classifying song reviews, the model leverages the information acquired from a source task or dataset, e.g., categorizing movie reviews \cite{ibm:transfer_learning}.

Furthermore, transfer learning has numerous advantages, including inexpensive processing costs, the flexibility to use huge pre-trained datasets, and broad application across a variety of tasks. First, it reduces the computing costs associated with developing models for new issues. By reusing pre-trained models or networks for other applications, users can save time, data, processing units, and other computing resources \cite{ibm:transfer_learning}. For instance, reaching the necessary learning rate may necessitate fewer epochs, thereby accelerating and simplifying model training procedures. Second, it considerably minimizes the challenges involved in collecting large datasets. Large language models, for example, need a massive quantity of training data to attain optimal performance. 

Finally, while this method aids in model optimization, it can also improve a model's generalizability. Since transfer learning involves retraining an existing model with a new dataset, the retrained model will be familiar with multiple datasets and may perform better on a wider variety of data compared to the original base model, which was trained on only one type of dataset \cite{ibm:transfer_learning}.

\subsection{Ensemble Learning}
\label{ensemble-learning}
Ensemble learning is a machine learning technique that combines two or more learners (such as neural networks or regression models) to generate better predictions \cite{murel2024ensemble}. In other words, an ensemble model makes predictions that are more accurate than those of a single model by combining multiple models \cite{murel2024ensemble}. There are two types of ensemble learning approaches in machine learning: parallel and sequential. In parallel techniques, each base learner is trained independently of the others, while sequential approaches train a new base learner to minimize errors made by the previous model trained in the preceding stage \cite{kunapuli2023ensemble}.

Parallel methods are further subdivided into homogeneous and heterogeneous methods \cite{murel2024ensemble}. To generate all of the component base learners in homogeneous parallel ensembles, the same base learning algorithm is used. In contrast, heterogeneous parallel ensembles use different algorithms to generate base learners \cite{zhou2012ensemble}. Additionally, some methods, such as stacking, can combine base learners into a final learner by separating the algorithms required to train an ensemble learner from the base learners \cite{murel2024ensemble}. Voting, specifically majority voting, is a typical method for condensing base learner predictions.

The final forecast produced by majority voting is based on the predictions made by the majority of learners, considering the predictions made by each base learner for a particular data instance \cite{murel2024ensemble}. For example, in a binary classification task, the majority vote utilizes the final prediction for a specific data instance, combining predictions from each base classifier \cite{murel2024ensemble}. A variation of this method is weighted majority voting, which prioritizes the predictions of some learners over others \cite{mienye2022survey}.

The most prevalent ensemble learning strategies are bagging, boosting, and stacking. The differences between sequential, parallel, homogeneous, and heterogeneous varieties of ensemble methods are best illustrated by combining them. A homogeneous parallel technique known as ``bootstrap aggregating'' is bagging \cite{murel2024ensemble}. To train multiple base learners with the same training method, it takes advantage of modified replicas of a particular training data set \cite{galar2011review}. Specifically, bagging trains multiple base learners by extracting several new datasets from a single initial training dataset using a process known as bootstrap resampling.

Stacking, also known as a layered generalization, is a heterogeneous parallel approach that exhibits meta-learning \cite{zhang2012ensemble}. Meta-learning is the process of training a meta-learner using the outputs of many base learners. Stacking trains several base learners from the same dataset using a unique training procedure for each learner \cite{murel2024ensemble}. Each base learner produces predictions on a previously unseen dataset. These initial model predictions are then combined and utilized to train the final model, which is the meta-model \cite{mienye2022survey}.

Boosting algorithms employ a sequential ensemble technique and it trains a learner on an initial dataset, and the resulting learner is often ineffective, misclassifying a large number of samples in the dataset. Like bagging, boosting takes instances from the initial dataset and creates a new dataset $(d2)$ \cite{murel2024ensemble}. However, unlike bagging, boosting prioritizes misclassified data instances from the initial model or learner. A new learner is then trained on this fresh dataset, $d2$ \cite{murel2024ensemble}. A third dataset $(d3)$ is subsequently created from $d1$ and $d2$, prioritizing the second learner's misclassified samples, as well as instances in which $d1$ and $d2$ disagree. This technique is repeated $n$ times to produce $n$ learners \cite{murel2024ensemble}. Boosting then integrates and weights all of the learners together to generate final predictions \cite{zhang2012ensemble}.

\section{Deep Learning}
\label{sec:deep-learning}

This section discusses frequently used deep learning models, including convolutional neural networks (CNN), recurrent neural networks (RNN), long short-term memory (LSTM), and generative adversarial networks (GANs). Furthermore, we cover backpropagation, a crucial training procedure for neural network optimization. In the following subsections, we provide a more detailed explanation of the operational principles of these designs.

\subsection{Convolutional Neural Network (CNN)}

Neural Network is a subset of machine learning. This network contains various nodes and layers; and there are lots of connections between nodes or neurons. These nodes are connected to each other associated with weight and threshold. The main structure of a neural network contains an input layer, one or more hidden layers, and the output layer. Once the output of each node is higher than the threshold value, then the information will be passed to the next layer. Convolutional neural networks are one of the important types of neural networks which are mainly used for classification, object detection and segmentation. 

In CNN, each convolutional layer can be followed by additional convolutional (Conv) layers, pooling layers, flatten, and fully connected (FC) layers. These layers are designed for better understanding and processing of data, features, and patterns. The input data passes through these layers to get precise output. During training, convolutional neural nets are learned to extract important and relevant features from data. Conv layers extract meaningful features like edges, textures and patterns from raw data with the aim of detection, classification, and prediction. Kernels are small filters that slide over these inputs and create feature maps. As the network deepens, the convolutional layers learn increasingly abstract representations of the input, making CNNs highly effective in recognizing complex structures within the data. We should be very careful about the complexity of the models as using deeper layers may cause complexity and slower processing. However, advanced CNN architectures such as ResNet and DenseNet have introduced innovations like residual connections and dense blocks, enabling the training of deeper networks without suffering from gradient degradation.

Pooling layers down sample these features to reduce dimensionality issues, ensuring that the most prominent features are retained while unnecessary details are discarded which allows the network to focus on key aspects of the data. Following the conv and pooling layers, these feature maps go through flatten layer, reshaping and converting the multidimensional tensor to 1D vector. Then to produce the final output, all previous neurons need to be connected, and this process is done in fully connected layer. The total number of neurons in the flatten layer equals the total number of components and elements in the previous layer. If the previous layer is a 3-dimensional tensor, then the number of neurons in flatten layer is the product of these dimensions. Additionally, the capacity of the network is dependent on the number of neurons in fully connected layer. As we have more neurons in FC layer, the model can learn more complex representations, but we should be careful about increasing the risk of computational cost and overfitting.

\subsection{Recurrent Neural Network (RNN)}

An RNN, or recurrent neural network, is a type of deep neural network trained on time series or sequential data to create a machine learning model that can produce sequential outputs in the form of predictions \cite{Stryker2024}. Like conventional neural networks, including feedforward and convolutional neural networks (CNNs), recurrent networks learn from training data. However, what sets them apart is their ability to use information from previous inputs to affect the present input and output \cite{Stryker2024}. They rely on the results of earlier parts in the sequence, while standard deep learning networks assume that inputs and outputs are independent. Although the outcome of a given sequence could also be influenced by future occurrences, unidirectional RNNs cannot incorporate these events into their predictions \cite{Stryker2024}.

To provide context for understanding RNNs, consider the expression ``feeling under the weather,'' which is often used to describe someone who is sick. The idiom must be expressed in that precise order to make sense. Thus, RNNs must consider each word's position within the idiom and use this knowledge to predict the next word in the sequence \cite{Stryker2024}. The words in the phrase ``feeling under the weather`` are not in any particular order; rather, they are part of a sequence \cite{Stryker2024}. Maintaining a hidden state at each time step allows the RNN to track the context. Moving the hidden state from one time step to the next creates a feedback loop. Information about prior inputs is stored in this hidden state, functioning as a kind of memory \cite{Stryker2024}. At each time step, the RNN processes the hidden state from the previous time step and the current input (such as a word in a phrase). As a result, it can remember past data points and apply that knowledge to affect the present output \cite{Stryker2024}.

Recurrent networks are also distinguished by their parameter sharing across all network layers \cite{Stryker2024}. Unlike feedforward networks, which have distinct weights for each node, RNNs use the same weight parameter throughout all layers. Nevertheless, these weights are still modified using gradient descent and backpropagation techniques to support reinforcement learning \cite{Stryker2024}. To find the gradients (or derivatives), RNNs employ forward propagation and backpropagation through time (BPTT) methods. This approach differs slightly from standard backpropagation because it is tailored to sequence data \cite{Stryker2024}. The fundamentals of BPTT are similar to those of conventional backpropagation, where the model learns by estimating errors from its input layer to its output layer \cite{Stryker2024}. These computations enable us to suitably adjust the model's parameters. The key difference with BPTT is that, while feedforward networks do not share parameters across layers, backpropagation through time (BPTT ) must accumulate errors at each time step \cite{Stryker2024}.

\subsection{Long Short-Term Memory (LSTM)}

The LSTM was introduced in \cite{hochreiter1997long} to address the vanishing gradient problem that arises with ordinary RNN and prevents the learning of long-term dependencies. The vanishing-gradient problem presents a situation in which the weights of RNN cease to change throughout the training process \cite{lindemann2021survey}. As a result, prioritizing current knowledge may result in the disregard of past occurrences. As a result, long-term dependencies are unable to be effectively learned. The LSTM is designed to regulate the whole information flow among neurons. To do this, a gating system is employed that governs the process of adding and removing information from an iteratively propagated cell state \cite{lindemann2021survey}. Thus, the forgetting process can be regulated, and a specified memory behavior is accomplished to simulate both short-term and long-term dependency.

In contrast to typical RNN, the neuron's output $h(t)$ is determined by the cell state $z(t)$, rather than the inputs $x(t)$ and prior outputs $h(t - 1)$ \cite{lindemann2021survey}. On the other hand, the LSTM gating mechanism determines the cell state. Vectors $f(t)$, the output of the forget gate, and $n(t)$, the output of the add gate, iteratively alter the cell state to govern memory behavior \cite{lindemann2021survey}. In the LSTM architecture, an inverse link between these two gates is employed to restrict memory capacity to some extent. As a result, each cycle adds and removes information from the cell state \cite{lindemann2021survey}.

This approach is driven by the fact that no memory is unlimited, and the human memory, as a role model, has a finite capacity. The output gate calculates the intended output by inferring the updated cell state. Hence, the most recent inputs do not always dominate the creation of output signals since the cell state includes a reduced and weighted representation of historical input information. This data is projected onto the output \cite{lindemann2021survey}. Therefore, the effect of important historical events, for example, is included in the output projection, and it is possible to ignore current inputs with low information density \cite{lindemann2021survey}. The gates are built using current inputs and past outputs. LSTM cells' adaptability has led to their use in a wide range of neural network architectures, each designed to address a particular set of issues \cite{lindemann2021survey}.

In other words, because LSTM networks are able to learn temporal associations and capture them in a low-dimensional state representation, they are destined to identify contextual anomalies \cite{lindemann2021survey}. These relationships deal with short- and long-term dependencies, as well as stationary and nonstationary dynamics. Multivariate time series and time-variant systems can be effectively modeled by LSTM networks \cite{lindemann2021survey}. As demonstrated in \cite{malhotra2015long}, which proposes a stacked LSTM architecture to detect abnormalities within time series data, LSTM-based techniques have demonstrated strong anomaly identification capacity.  An LSTM network is used in \cite{Ergen_2020} to forecast regular system dynamics, and a support vector machine is used to classify anomalies in order to provide a self-learning detection mechanism. In \cite{bontemps2016collective}, a method for detecting collective abnormalities using LSTM networks is proposed, which involves analyzing numerous one-step forward prediction errors. A real-time detection technique is implemented in \cite{lee2020rere} using two LSTM networks, one of which models short-term properties and the other of which controls the detection by means of long-term thresholds.

\subsection{Generative Adversarial Networks (GANs)}

Goodfellow initially introduced GANs in 2014 \cite{goodfellow2014generative}. One simple concept is to have two models combat each other \cite{liu2020survey}. One model is generator $G$, while the other is discriminator $D$. Then it must provide the $G$ with a random noise $z$ that follows the prior probability, and the $G$ will output the data as $G(z)$. Finally, the discriminator is updated using the $G(z)$ and $P(\text{data} \, x)$ values \cite{liu2020survey}. Following training, the $D$ network determines whether the input data is real or created by the generator. The $D$ improves its discriminating capacity through continual learning, whereas the $G$ makes its data more realistic through continuous learning. To trick the discriminator, $G$ and $D$ combat each other, strengthening their abilities and eventually reaching a stable state. The discriminator is unable to distinguish the generator's data, resulting in falsification \cite{liu2020survey}.

There are some issues with the original GANs \cite{liu2020survey}. First, there is instability in the training. Ensuring the synergy between $G$ and $D$ is a challenge. Second, there is little variation in the fake photos. Lastly, there isn't a consistent effective criterion for the resulting image's quality. As a result, several GAN enhancements have lately been suggested. As an example, DCGAN (deep convolutional GAN) was proposed by Radford et al. \cite{radford2015unsupervised}. The network structure of the original GANs is primarily enhanced by DCGAN. It substitutes two convolutional neural networks for the generator and discriminator, increasing network stability but failing to address the core issue \cite{liu2020survey}.

Arjovsky introduced WGAN (Wasserstein GAN) \cite{arjovsky2017wasserstein}, which substituted the Wasserstein distance (EM distance) for the $JS$ distance. The instability problem can be substantially resolved by WGAN. It was hypothesized by Gulrajani et al. \cite{gulrajani2017improved} that WGAN-GP can enhance WGAN. Additionally, EBGAN \cite{zhao2016energy} and BEGAN \cite{berthelot2017began} are capable of producing incredibly realistic face images. PGGAN (Pro-GAN), PGGAN, and StyleGAN \cite{karras2019style} were suggested by TeroKarra et al. \cite{karras2017progressive} and employed a step-by-step process to produce high-resolution, high-quality pictures. Some GANs may translate images from one format to another and change their style, such as CycleGAN \cite{zhu2017unpaired} and StarGAN \cite{choi2018stargan}.

\subsection{Transformer Architecture}

The transformer model was first introduced in \cite{vaswani2017attention} for machine translation work. Since then, many other models have been created to meet a range of problems in many fields, drawing inspiration from the original transformer model \cite{vaswani2017attention}. The task and performance of transformer-based models can therefore differ according to the particular architecture used. However, self-attention is a crucial and frequently utilized part of transformer models, and it is necessary for their functionality \cite{vaswani2017attention}. The self-attention mechanism and multi-head attention, which normally comprise the architecture's first learning layer, are used by all transformer-based models \cite{vaswani2017attention}.

The attention mechanism has become more popular because of its ability to focus on important information \cite{ibm_transformer_model}. Certain visual regions were discovered to be more relevant than others while processing images. As a result, the attention mechanism was proposed as a novel approach to computer vision tasks, to emphasize key components based on their contextual relevance inside the application. When applied to computer vision, this technique produced remarkable results, promoting its widespread use in a variety of other domains, including language processing.

To overcome the shortcomings of other neural networks, such as recurrent neural networks (RNN) in recording long-range dependencies in sequences, particularly in language translation tasks, a unique attention-based neural network known as ``Transformer'' was presented in 2017 \cite{vaswani2017attention} \cite{transformer2024}. By lowering the dependence on outside data and better capturing local features, the transformer model's self-attention mechanism enhanced the attention mechanism's performance \cite{vaswani2017attention}. The ``Scaled Dot Product Attention'' in the original transformer design is used to execute the attention approach \cite{vaswani2017attention}. It is based on three main parameter matrices: query $(Q)$, key $(K)$, and value $(V)$ \cite{vaswani2017attention}. Every matrix in the series has an encoded representation of every input \cite{vaswani2017attention}. The scaled dot product attention function can be calculated mathematically in the following way:

\begin{equation}
    Attention(Q, K, V) = softmax \left( \frac{QK^T}{\sqrt{d_k}} \right) V
\end{equation}

The value vectors are represented by matrix $V$, whereas the query and key vectors, with a dimension of $dk$, are represented by matrices $Q$ and $K$, respectively.

To extract the maximal interdependence between various segments in the input sequence, the multi-head attention module must apply the scaled dot-product attention function simultaneously \cite{vaswani2017attention}. The attention mechanism is carried out by each head, represented by the letter $k$, using its unique learnable weights, $W^{kQ}$, $W^{kK}$ and $W^{kV}$. According to Vaswani et al. \cite{vaswani2017attention}, the attention outputs computed by each head are then concatenated and linearly transformed into a single matrix with the anticipated dimension.

The neural network can learn and capture many features of the sequential input data more easily when multi-head attention is used \cite{vaswani2017attention}. As a result, this improves the representation of the input contexts by combining data from various attention mechanism aspects within a given range—which may be brief or long. Better network performance is the outcome of this strategy since it permits the attention mechanism to operate cooperatively \cite{vaswani2017attention}. The auto-regressive sequence transduction model served as the foundation for the development of the original transformer architecture, which consisted of the encoder and detector modules \cite{vaswani2017attention}. Each module integrates the attention mechanism over multiple layers. Specifically, the transformer architecture's many concurrent executions of the attention mechanism account for the existence of multiple ``Attention Heads'' \cite{vaswani2017attention}.

\subsection{Backpropagation}

\begin{figure}[ht]
\centering
\includegraphics[width=0.9\linewidth]{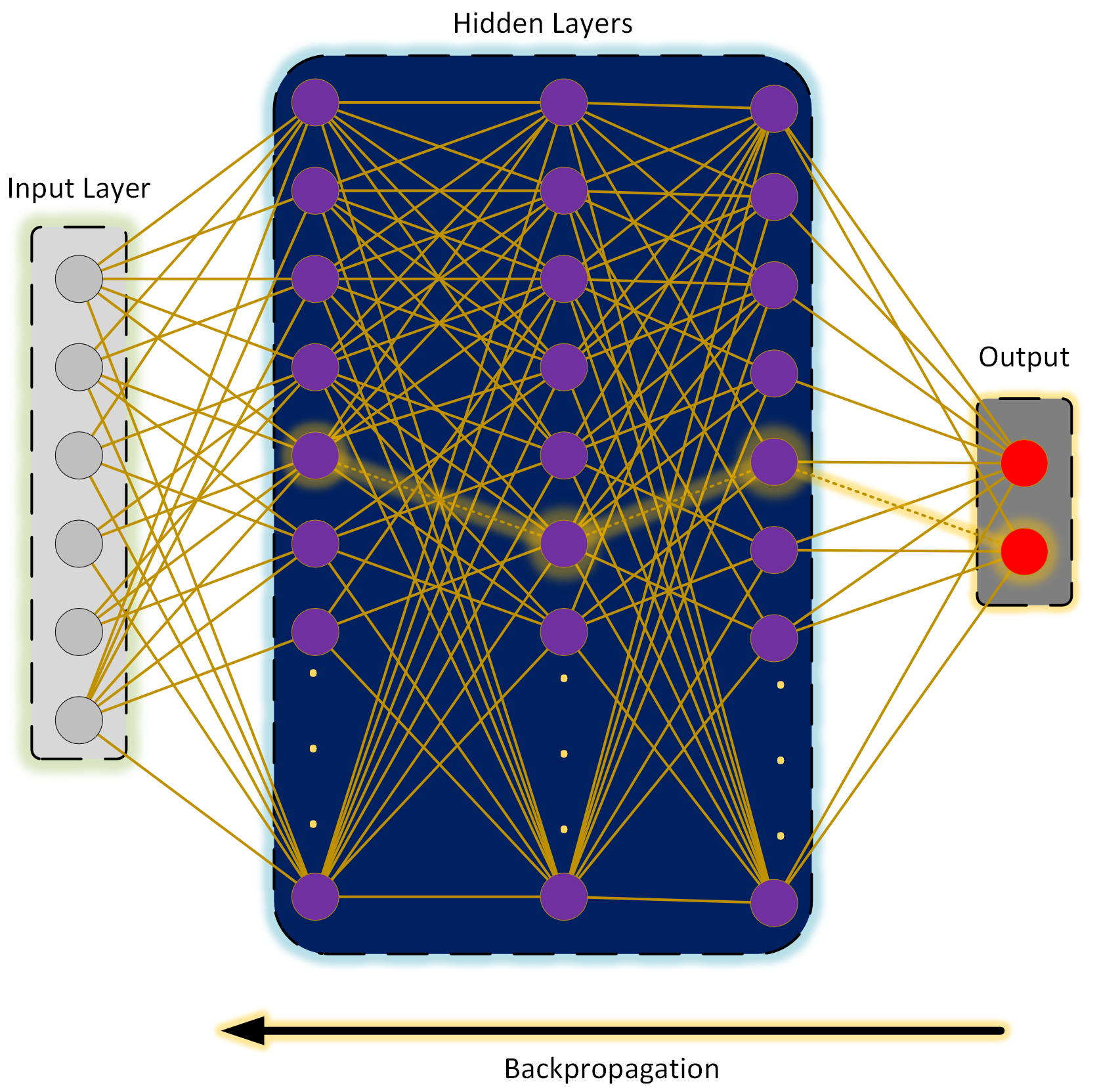}
\caption{ Visualization of the Backpropagation algorithm.}
\label{fig:back-propagation}
\end{figure}

The backpropagation algorithm calculates the gradient of a neural network's loss function in terms of weights and biases \cite{stergiou1996neural}. It uses the chain rule to compute the partial derivatives of the loss function with respect to each weight and bias, starting from the output layer and working backward to the input layer (see Fig.
\ref{fig:back-propagation}).

Consider a neural network with one node per layer. The following equations \cite{stergiou1996neural} calculate the cost, weighted sum, and activated weighted sum:

\begin{align}
z^L &= w^L a^{L-1} + b^L \\
a^L &= \sigma(z^L) \\
C &= (a^L - y)^2
\end{align}

where \(z^L\) represents the current layer's weighted sum, \(w^L\) denotes the weights linking the current layer to the preceding layer, \(a^{L-1}\) represents the neuron's activation in the previous layer, \(b^L\) indicates the bias for the present layer, and \(C\) represents the Mean Squared Error (MSE) loss function, calculating the difference between the predicted output and the target value.

The derivative for the weight between the hidden and output layers is given by \cite{stergiou1996neural}: 

\begin{equation}
\frac{\partial C_k}{\partial W^L} = a^{L-1} \cdot \sigma'(z^L) \cdot 2(a^L - y)
\end{equation}

Here, \(\frac{\partial C_k}{\partial W^L}\) denotes the partial derivative of the loss function \(C_k\) with regard to weight \(W^L\), assessing how the weight reduction responds to variations in that weight. This process iterates for each weight, bias, and activation number in the system. The average derivative \cite{stergiou1996neural} across data points is computed as:


\begin{equation}
\frac{\partial C}{\partial W^L} = \frac{1}{n} \sum_{k=0}^{n-1} \frac{\partial C_k}{\partial W^L}
\end{equation}

Neural networks typically include numerous inputs, outputs, and nodes, making the calculations complex. With computational assistance, the neural network performs these calculations repeatedly until a minimum is attained.

\section{Hybrid Models}
\label{sec:hybrid-models}
Although CNN itself is a powerful classification model but some believe that using hybrid CNN-ML models provides us with better results. As it is obvious, ConvNets can provide high dimensional vectors, and ML models like SVM can handle these complex decision boundaries well.  So, in the hybrid-type models, CNN can be used to extract the meaningful features and during the training, these features are learned for some specific tasks (see Fig. \ref{fig:knn-step}). The features represent the input data (e.g. images). Each image is now a vector of learned features, and these features are positioned in the last convolutional layer before the fully connected layers. 
Here is an interesting question: will the features learned in the last convolutional layer—just before the fully connected layer—be fed directly into the SVM?
The layer that will be fed to SVM could either be the flatten layer which is the 1D representation of the features maps or the FC layer. 1D flatten layer represents the learned features obtained from last conv layer. If there will be a max pooling layer right before the flatten layer, then the number of neurons in flatten layer would be equal to the product of the dimensions of this max pooling layer. Thus, this flatten layer will be fed into a ML model like SVM as an input for further classification. The SVM is trained using features extracted from the CNN. These features, learned during CNN training, serve as input to the SVM, which then learns how to perform classification. Since this is a hybrid model, an important consideration is whether the features are learned before or during SVM training. In fact, the features are already learned during the CNN training phase. The SVM  is trained separately using these extracted features. Therefore, hybrid models like this involve two training stages: first, the CNN is trained on the input data to extract features, and then the SVM is trained on those features to perform classification and prediction.

\section{ Explainable Artificial Intelligence (XAI)}
\label{sec:xai}

We can think of machine learning (ML) models as a black box. However, explainable artificial intelligence (XAI) techniques have made this process easier to understand \cite{salih2024perspective}. With the goal of increasing end users' trust in machine learning models and increasing transparency, these techniques aid in explaining how the model operates \cite{salih2024perspective}. Two popular XAI techniques are Shapley Additive Explanations (SHAP) and Local Interpretable Model Agnostic Explanations (LIME), especially when working with tabular data \cite{salih2024perspective}.
Based on game theory, SHAP is an XAI technique that attempts to explain any model by treating all features (or predictors) as players and the model outcome as the payout. In other words, SHAP can explain the role of characteristics for both a particular instance and for all instances \cite{salih2024perspective}. It also offers both local and global explanations. It has been extensively used to explain model results both locally and globally across a variety of domains \cite{salih2024perspective}. But before using SHAP, end users need to be aware of a few key considerations. To begin with, SHAP is a model-dependent technique. This indicates that different explainability scores may result from the SHAP conclusion depending on the ML model applied to the classification or regression task \cite{salih2024perspective}. As a result, the top features found by SHAP may vary throughout ML models when many models are used for the same task with identical data.
Another XAI technique called LIME seeks to describe the model's local operation for a particular instance \cite{salih2024perspective}. In order to achieve this, it approximates the complex model's predictions by converting it into a smaller, more understandable model that describes the model's behavior in that specific instance \cite{salih2024perspective}. LIME turns any model into a linear local model and reports the coefficient values, which represent the weights of the features in the model.
However, as LIME is based on a linear approximation, it might miss nonlinear linkages present in the original model, which could result in models with substantial nonlinear interactions having inadequate explanations \cite{salih2024perspective}. Additionally, LIME is model-dependent, which means that even though the task and dataset are the same, its outcomes could change based on the particular model being explained \cite{salih2024perspective}.

\section{Fooling Learning Algorithms into Misclassification}

\label{sec:fooling-learning}

\begin{figure}[ht]
\centering
\includegraphics[width=0.6\linewidth]{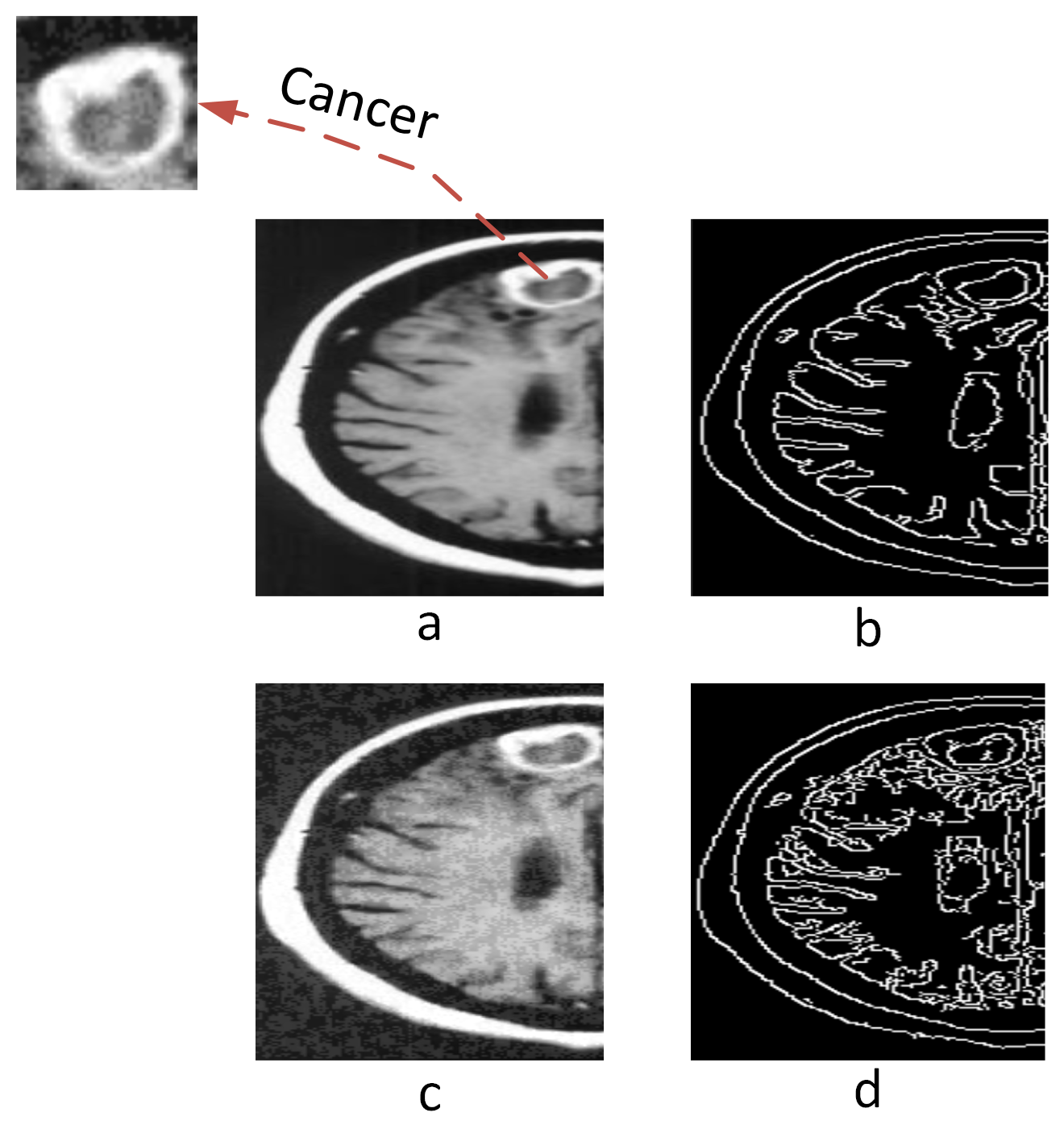}
\caption{Effect of FGSM adversarial noise on the image. (a) clean image, (b) clean edge, (c) noisy image, (d) noisy edge.}
\label{fig:missclassi-ml}
\end{figure}

An adversarial attack is a term used to describe a system designed to fool learning-based models into making mistakes (misclassifying an image). They exploit weaknesses in the models and take advantage of properties such as shortcutting in order to subtly alter an image just enough to make the model classify an image differently. Modern machine learning models are susceptible to these attacks, as they have inherent limitations and vulnerabilities, ranging from the data they are trained on to their parameters and lack of transparency. Through these multiple limitations, adversarial attacks can be created with surprising ease, and go on to become very effective and discreet, as they are often not detectable by simply looking at the images. Examples include attack methods such as Projected Gradient Descent (PGD) and Fast Gradient Sign Method (FGSM). Even high-performing state-of-the-art models can be very susceptible to attack. It is extremely important to study adversarial noise removal and work to eliminate malicious attempts by developing robust noise-informed learning algorithms, as successful attacks could lead to disastrous consequences in many fields, such as healthcare or autonomous driving. For example, as can be seen from Fig. \ref{fig:missclassi-ml}, adversarial noise affected the cancer edges by adding more pixels around them, which may lead to misclassifying the cancerous image into non-cancerous. The dataset is available in \cite{Br35H2020}.  

\begin{figure*}[ht]
\centering
\includegraphics[width=0.8\linewidth]{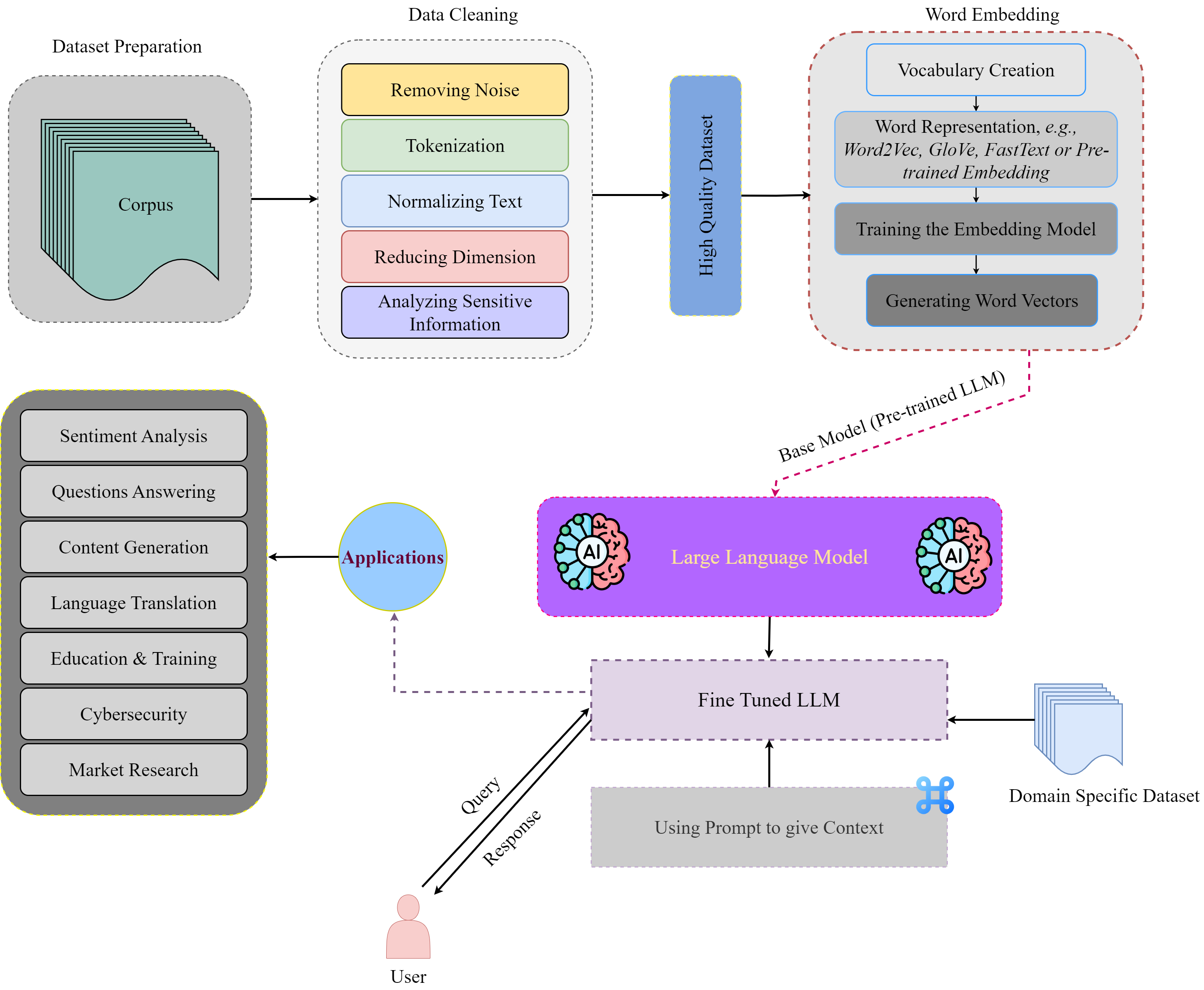}
\caption{Overview of fine-tuning Large Language Models (LLMs). This process includes dataset preparation, word embedding, and adapting a pre-trained model on domain-specific data to enable applications like sentiment analysis, question answering, and content generation. Users interact with the LLM through prompts to receive relevant responses.}
\label{fig:llm-intg-arch}
\end{figure*}

\section{LLM Integration}
\label{sec:llm-integration}

In this section, we will review several downstream tasks where LLMs are effectively utilized and examine key parameters that provide deeper insights into their performance and applications.

\subsection{How LLMs Power Real-World Applications?}
Large language models (LLMs) are a type of foundation model trained on vast amounts of data, enabling them to generate and understand various forms of content, including natural language, for diverse applications \cite{ibmLLM2024}. Today, researchers are leveraging LLMs to develop a range of significant applications, as illustrated in Fig. \ref{fig:llm-intg-arch}. In this figure, we demonstrate how fine-tuning LLMs can create applications such as sentiment analysis, question answering, content generation, language translation, education and training, cybersecurity, and market research.
The differences between traditional machine learning (ML), deep learning (DL), and LLMs are substantial. In simple terms, LLMs are significantly larger in terms of parameters (often billions or trillions), enabling them to generate vast amounts of content and capture complex contextual relationships in user queries.

As shown in Fig. \ref{fig:llm-intg-arch}, the fine-tuning process for LLMs involves five key steps. The process begins by taking a text corpus and preprocessing it since high-quality datasets are essential for robust performance in any ML model, including language models as well. After data cleaning, the model generates word embeddings using techniques like Word2Vec, GloVe, FastText, or other pre-trained embedding models. Next, a base LLM, such as GPT or LLaMA, needs to be selected. Since training an LLM from scratch may not be feasible, fine-tuning is employed to quickly achieve desired outputs. Once the base model is selected, fine-tuning begins by providing prompts and instructions that guide the LLM to generate personalized responses. In addition, a domain-specific external dataset is also used to further enhance its capabilities. Finally, users can ask questions and receive responses

Table \ref{tab:llm_overview} provides an overview of LLM applications across healthcare, education, security, language processing, business, and content creation. In the healthcare sector, these models have been adopted to automate administrative tasks and develop Question answering (QA) systems. Hospitals, which traditionally require significant manpower for administrative duties, are now automating these processes using LLMs like GPT and Llama. Additionally, Question-answering (QA) frameworks allow patients to ask questions and receive automated responses without waiting for a human agent.

In terms of security, LLMs have been applied to vulnerability detection because of their billions of parameters and extensive training on vast datasets. Therefore, these models can quickly identify vulnerabilities by leveraging their deep understanding of the data.

Language processing has seen a surge in sentiment analysis, allowing businesses to analyze user sentiments from large text corpora. This capability is particularly useful in business applications where understanding user reactions to products can inform stakeholders and lead to improved product quality. Moreover, LLMs have also revolutionized the business sector, with applications like chatbots and stock market predictions. These advancements offer significant advantages in today's data-driven era.

In content creation, LLMs have proven to be highly beneficial. Users can generate content based on specific use cases, such as writing movie scripts or creating stories—tasks that are typically time-consuming for humans. Hence, LLMs are valuable for content analysis, especially in tasks that heavily rely on processing large amounts of content. This technology has tremendous potential in today's modern world.

Looking at Table \ref{tab:llm_overview}, we can also see that researchers have been very interested in fine-tuning LLMs; very few studies have not addressed this topic. Furthermore, deployment was not the main focus of the previous research listed in Table \ref{tab:llm_overview}, which meant that the models were maintained in the testing stage, where they are still undergoing constant improvement. In addition, a group of researchers \cite{neupane2024questionsinsightfulanswersbuilding} concentrates on both deployment and fine-tuning because an end-to-end solution was considered while implementing a tailored chatbot to access university resources.

\subsection{LLM Reasoning}

LLM reasoning is the ability of large language models to imitate human-like cognitive processes when addressing problems requiring multi-step logic, deduction, or inference. LLMs that can solve mathematical problems, complete code, and conduct multi-step logical inference show remarkable reasoning abilities.  However, models like as GPT-4 and LLaMA3 continue to struggle with accuracy in complicated, multi-step reasoning, where surface-level correctness does not ensure logical consistency.

To address this, recent work by \textit{Zhenwen et al.} \cite{liang2024improving} proposes a verification framework to enhance inference-time reasoning through verifiers such as \textsc{Math-Rev} and \textsc{Code-Rev}. These are trained using SimPO—a reference-free preference optimization technique—to distinguish correct from incorrect outputs sampled from various LLMs. The authors also introduce \textsc{CoTnPoT}, a hybrid technique that transforms natural language reasoning steps \( S_{\text{CoT}} \) into executable programs \( S_{\text{PoT}} \):
\begin{equation}
S_{\text{PoT}} = \text{CoderLLM}(Q, S_{\text{CoT}})
\end{equation}

where \(Q \) is the original problem prompt, \(S_{\text{CoT}} \) is the chain-of-thought solution in natural language, and \(\text{CoderLLM}(\cdot) \) is a coder model that turns reasoning stages into executable code-like representations for validation.

During inference, solutions are scored using Gumbel-Softmax weighted voting as follows:

\begin{equation}
y_i = \frac{\exp\left(\frac{\log(\pi_i)}{\tau}\right)}{\sum_{j=1}^{k} \exp\left(\frac{\log(\pi_j)}{\tau}\right)}
\end{equation}

where \(\pi_i \) is the unnormalized score awarded by the verifier to the \(i\)-th solution, \(\tau \) is the temperature parameter determining the sharpness of the distribution, and \(y_i \) is the normalized weight utilized for final decision-making. 

However, this approach improves reasoning accuracy on datasets like GSM8k \cite{cobbe2021gsm8k} and MathBench \cite{liu2024mathbench}, outperforming even GPT-4o when paired with strong backbone models. Building upon the need for standardized evaluation, \textit{Hao et al.} \cite{hao2024llmreasoners} propose AutoRace, an automated reasoning chain evaluator that identifies common reasoning errors without relying on handcrafted prompts. It defines a binary metric \( s(z) \in \{0,1\} \) based on whether an LLM-generated reasoning chain \( z \) leads to a correct answer \( y \) matching the reference \( y_r \). Alongside this, the authors introduce LLM Reasoners, a modular library unifying reasoning algorithms via a search-based formulation, which is defined as follows:

\begin{equation}
\arg\max_{a_0, \ldots, a_T} \sum_{t=0}^{T} r(s_t, a_t), \quad s_t \sim T(\cdot | s_{t-1}, a_{t-1})
\end{equation}

where \( a_t \) is the action performed at step \( t \), \( s_t \) is the reasoning state that results, \( r(s_t, a_t) \) is the reward function that assesses the quality of each reasoning step, and \( T(\cdot | s_{t-1}, a_{t-1}) \) is the world model that defines the state transitions based on earlier actions. 

This framework supports consistent comparison of CoT, Tree-of-Thought (ToT), and Reasoning via Action Planning (RAP), with RAP showing superior performance by combining explicit world models with reward-guided search (e.g., Monte Carlo Tree Search). Complementary to these structural and evaluative advances, \textit{Han et al.} \cite{han2024token} address the inefficiency of CoT-style reasoning, which often incurs excessive token usage. They introduce TALE—a Token-budget-Aware LLM reasoning framework—to reduce costs while preserving performance. TALE includes two methods: TALE-EP, which uses zero-shot token budget estimation to create concise prompts, and TALE-PT, which trains LLMs to internalize token-efficient reasoning via Supervised Fine-Tuning (SFT) or Direct Preference Optimization (DPO). These are trained using objectives and we can represent them mathematically as follows:

\begin{equation}
\mathcal{L}_{\text{CE}}(\theta) = -\frac{1}{N} \sum_{i=1}^{N} \sum_{t=1}^{T_i} \log P(y_{i,t} \mid y_{i,<t}, x_i)
\end{equation}
where \( \mathcal{L}_{\text{CE}} \) denotes the cross-entropy loss over the training set, \( y_{i,t} \) is the target token at time step \( t \), \( y_{i,<t} \) is the sequence of previous tokens, and \( x_i \) is the input question for the \( i \)-th example.

\begin{equation}
\mathcal{L}_{\text{DPO}}(\theta) = -\frac{1}{N} \sum_{i=1}^{N} \log \frac{\exp(s(y_i, x_i))}{\exp(s(y_i, x_i)) + \exp(s(y'_i, x_i))}
\end{equation}
where \( s(y, x) \) represents the log-likelihood score of a candidate output \( y \) given input \( x \), and \( y_i \succ y'_i \) indicates that the model should prefer the token-efficient output \( y_i \) over the baseline output \( y'_i \). 

This preference-based loss function encourages alignment with concise and accurate responses. Results on benchmarks: GSM8k \cite{cobbe2021gsm8k} and MathBench \cite{liu2024mathbench} reveal that TALE achieves up to 68\% reduction in token usage and 59\% in cost, with minimal accuracy loss. Together, these efforts advance the field of LLM reasoning by tackling performance, evaluation, and efficiency, laying the groundwork for more accurate, interpretable, and cost-effective reasoning systems.

\begin{table*}[!ht]
\centering
\scriptsize  

\caption{Overview of LLM Applications in Healthcare, Education, Security, Language Processing, Business, and Content Creation} 

\renewcommand{\arraystretch}{1.5} 
\definecolor{headergreen}{RGB}{0,128,0}
\definecolor{skyblue}{RGB}{135,206,235}
\definecolor{coral}{RGB}{255,127,80}
\definecolor{goldenrod}{RGB}{218,165,32}
\definecolor{slategray}{RGB}{112,128,144}
\definecolor{tomato}{RGB}{255,99,71}

\definecolor{lightslategray}{RGB}{138,143,150}
\definecolor{darkslategray}{RGB}{47,79,79}
\definecolor{slategraydark}{RGB}{85, 92, 99}
\definecolor{slategraylight}{RGB}{169, 186, 195}
\definecolor{slateblue}{RGB}{106,90,205}

\begin{tabular}{>{\centering\arraybackslash}p{3cm}>{\centering\arraybackslash}p{4cm}>{\centering\arraybackslash}p{4cm}>{\centering\arraybackslash}p{2cm}>{\centering\arraybackslash}p{2cm}}
\cline{1-5}
\cline{1-5}
\cellcolor{slategray}\textcolor{white}{\textbf{Application}} & 
\cellcolor{slategray}\textcolor{white}{\textbf{Methods/Context}} & 
\cellcolor{slategray}\textcolor{white}{\textbf{Improvements}} & 
\cellcolor{slategray}\textcolor{white}{\textbf{Fine-tuning}} & 
\cellcolor{slategray}\textcolor{white}{\textbf{Deployment}} \\
\cline{1-5}

\multirow{4}{*}{Healthcare} 
& LLM-based multi-agent framework \cite{10527275}  & Automates administrative tasks & \checkmark & \texttimes \\ 

& HealthQ Framework \cite{wang2024healthqunveilingquestioningcapabilities} & Enhances questioning capabilities in LLM healthcare chains & \checkmark & \texttimes \\ 
\hline
\vspace{1.7pt}
\multirow{3}{*}{Education} 
& Retrieval Augmented Generation \cite{neupane2024questionsinsightfulanswersbuilding} & Enhances user access to campus resources & \checkmark & \checkmark \\ 

& Automatic plan generation using LLMs (T5, GPT-3.5) \cite{goslen2024llm}  & Enhances adaptive scaffolding in game-based learning & \checkmark & \texttimes \\ 
\hline
\vspace{1.7pt}
\multirow{2}{*}{Security} 

& Vulnerability detection using GPT-3.5-Turbo \cite{10456393} & Enhances software security through continuous monitoring & \checkmark & \texttimes \\ 
& Vulnerability detection using LLM \cite{lu2024grace} & Enhances vulnerability detection efficiency & \checkmark & \texttimes \\ 
\hline
\vspace{1.7pt}
\multirow{2}{*}{Language Processing} 
& Multi-Agent framework for financial sentiment analysis \cite{xing2024designing} & Enhances accuracy by leveraging discussions among multiple LLM agents & \texttimes & \texttimes \\ 
& DeepExtract: Semantic-driven hierarchical analysis \cite{onan2024deepextract} & Enhances accuracy and efficiency & \checkmark & \texttimes \\ 
\hline
\vspace{1.7pt}
\multirow{2}{*}{Business} 

& Stock market trend prediction \cite{swamy2023llm} & Achieves high accuracy in predictions & \checkmark & \texttimes \\ 
& Business analytics chatbot based on GPT-4 \cite{ngo2024babot} & Enhances decision-making & \checkmark & \texttimes \\ 
\hline
\vspace{1.7pt}
\multirow{3}{*}{Content Creation} 
& Rambler (Dictation Tool) \cite{lin2024rambler} & Enhances iterative revisions and user control & \texttimes & \texttimes \\ 
& LLM-assisted content analysis \cite{chew2023llmassistedcontentanalysisusing} & Reduces time for deductive coding, aligns with human accuracy & \checkmark & \texttimes \\ 
\hline
\end{tabular}
\label{tab:llm_overview}
\end{table*}

\section{Real-world Applications}
\label{sec:real-world-applications}

In this section, we explore real-world applications of some learning-based algorithms, as well as their methodologies being used in various fields.

\begin{figure*}[ht]
\centering
\includegraphics[width=\linewidth]{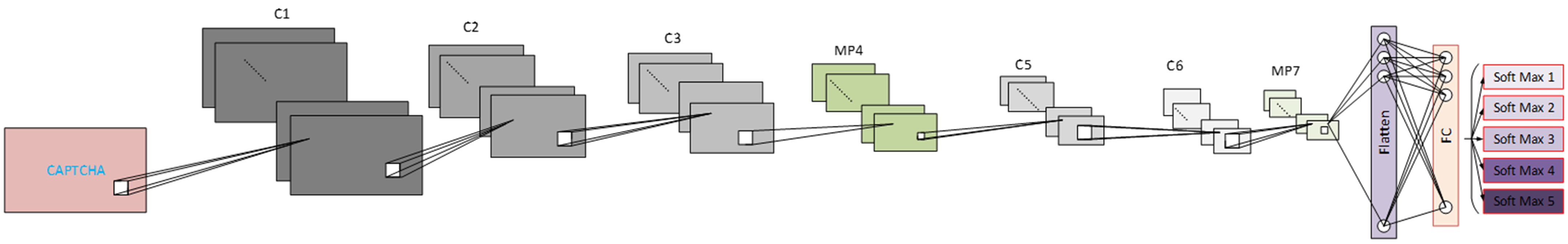}
\caption{Visualization of the CNN architecture for text-based CAPTCHA recognition. The CAPTCHA contains 5 characters.}
\label{fig:captcha-cnn-arch}
\end{figure*}

The CNN architecture in Fig. \ref{fig:captcha-cnn-arch} (which can be considered as an improved version of LeNet-5) consists of convolutional layers, max-pooling layers, flatten layer, and a fully connected layer with dropout. The convolutional and max pooling layers have specified kernel size and stride followed by an activation function. In the end, we have \(K\) softmax function in which \(K\) is the number of characters in each image. The loss function can be the softmax loss. SGD algorithm can be used for training the net with a specified learning rate. 

This CNN is going to recognize the CAPTCHAs containing English alphabetical letters and digit numbers which may also be available in these CAPTCHAs \cite{stark2015captcha}. As it is clear, we have 10-digit numbers (0-9), 26 small alphabetical letters (a, b, c, …, z), and 26 capital alphabetical letters (A, B, C, …, Z) which altogether there will be 62 characters (see the equation below). So, in the output layer, any character will be predicted by 62 neurons \cite{stark2015captcha}. If our CAPTCHA contains 4 characters, then \(4 \times 62 = 248\) neurons will be dealing with predicting the characters.


\begin{equation}
T(i)= \left\{ 
  \begin{array}{ll}
    1,2,3,\ldots,26 & \text{if } i = a-z \\
    27,28,29,\ldots,52 & \text{if } i = A-Z \\
    53,54,55,\ldots,62 & \text{if } i = 0-9
  \end{array}
\right. 
\end{equation} 

where \( i \in \{a-z, A-Z, 0-9\} \).

Some of the CNNs mostly dealt with single characters. The input image is a character in the square frame like \(30 \times 30\), but as can be seen from the architecture in Fig. \ref{fig:captcha-cnn-arch}, the input can be the whole CAPTCHA image. Based on the above equation, this CNN recognizes digits as well as small and capital characters at the same time.

In this CNN, we can obtain the output of each layer using the below equation:

\begin{equation}
O = \frac{i - k + 2p}{s} + 1
\label{eq:output}
\end{equation} 

where \(O\) is the output of each layer, \(i\) is the input, \(k\) denotes the kernel, \(p\) is padding, and \(s\) is stride.

There are various types of convolutional neural networks such as ResNet50 \cite{he2016deep}, DenseNet121 \cite{huang2017densely}, ALEXNET \cite{krizhevsky2017imagenet}, LeNet-5 \cite{lecun1998gradient}, VGG-16 \cite{simonyan2014very}, etc. CNNs have various applications and can be used to solve real world problems. In \cite{ucckun2020direction}, CNN has been used for direction finding which is based on learning signals reaching to the antenna. In another study, the authors utilized CNN for computer network enhancement optimization algorithm \cite{li2021computer}. Xin et al. proposed a new framework for complex network classification based on convolutional neural network \cite{xin2020complex}.

In another study conducted by Moore et al., CNN has been used for spike transformation with a generalized linear model \cite{moore2020validation}. Deep CNN has been used in \cite{almakky2019deep} for biomedical literature text localization. CNN has also been utilized for voice activity detection \cite{li2021tibetan} and face image recognition \cite{lou2020face}. CNN-based optimal time window derivation of human activity detection has been proposed in \cite{lee2019optimal}. It has also been used in medical applications such as brain MR image segmentation \cite{lahoti2021whole}, brain tumor classification \cite{sultan2019multi}, and tumor detection \cite{li2019brain}. It has also been applied for network intrusion detection \cite{li2020research, zheng2021network}. Fashion class classification \cite{stephen2019multiple}, lung cancer classification \cite{matsubara2019convolutional}, channel quantization \cite{cho2020per}, object grasping \cite{revathi2019object}, masked neural style transfer \cite{handa2018masked}, cherry image recognition \cite{zhenmin2022cherry}, speech emotion recognition \cite{wani2020speech}, finger vein recognition \cite{zhang2020research}, abnormal network traffic recognition \cite{liu2022abnormal}, fault detection and diagnosis \cite{tingting2020improved}, speech recognition \cite{yang2020speech}, gesture recognition \cite{zhiqi2020gesture}, potato leaf health detection \cite{9181312}, whistle recognition \cite{9622085}, image super-resolution \cite{savvin2020algorithm}, visibility estimation of CCTV images \cite{giyenko2018application}, and breast histopathological image classification \cite{gao2022convolutional} are a few applications of convolutional neural network.

\begin{figure*}[ht]
\centering
\includegraphics[width=0.8\textwidth]{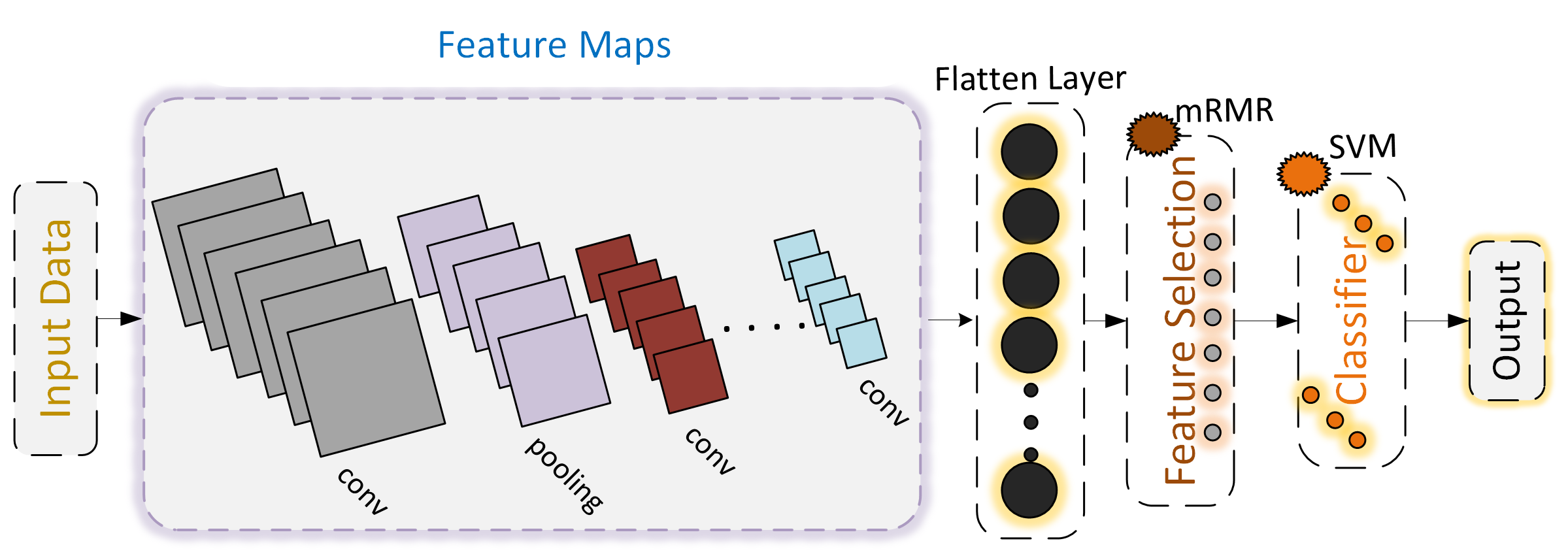}
\caption{Visualization of the architecture of hybrid CNN-SVM model for classification task.}
\label{fig:knn-step}
\end{figure*}

Additionally, CNN combined with ML models such as SVM can be used for various applications such as COVID-19 detection \cite{addeh2021hybrid}, text-based CAPTCHA classification, control chart pattern recognition, and more. For example, in the hybrid model, CNN can be used to extract the features, followed by feature selection algorithms such as Minimum Redundancy Maximum Relevance (mRMR) to select the most important ones. These important features can be fed to ML (SVM) to learn the classification task (see Fig. \ref{fig:knn-step}).


In addition to these, ML has significant applications in the healthcare sector, as demonstrated in \cite{siddiqui2022analysing}, where researchers identified COVID-19 mortality risk factors using ML algorithms. This serves as a reminder of the effectiveness of this technology in crucial areas like healthcare. Additionally, ML methods have been employed to develop next-generation eco-friendly travel systems aimed at reducing carbon emissions \cite{hossain2022hybrid}. Furthermore, large language model (LLM) technology, a state-of-the-art approach, has been used to create a chatbot system for assisting prospective and current students at Mississippi State University. However, ML has numerous other applications, making our lives significantly easier.

\section{Future of Learning Algorithms} 

\label{sec:future-of-learning}
The future of neural networks and learning algorithms as a whole is incredibly exciting, and with the wealth of different types of model, and the different goals that they strive to achieve, there are many possibilities on how to improve next. Currently, many of the top neural networks, mainly models such as transformers, currently rely on incredibly large amounts of data in order to achieve their state-of-the-art results. This is currently leading the community to a trajectory of more input data and more processing power, an unsustainable route. Neural networks began as a way machines could mimic human thought, and humans think through extremely efficient pathways, compartmentalizing and focusing on specific points of information quickly, which current top models do not follow. In the future, a different framework of model, which more closely aligns with the neural pathways of organic life, could provide state-of-the-art results with only a small fraction of the data, more akin to how humans interact with information. These future models could solve the data problem, while also opening further pathways for better unsupervised learning.

Looking ahead to the evolution of learning algorithms, we anticipate a transition toward flexible and adaptable neural networks that can effortlessly process a diverse array of inputs. At present, deep neural networks are tailored for specific tasks and excel only when dealing with limited types of data inputs. The next challenge in this field involves creating models that are not just specialized but highly adaptable—a framework capable of intelligently analyzing and forecasting outcomes from a wide spectrum of data formats. Central to this concept is a framework that consolidates diverse input types into a standardized format, similar to how barcodes provide a universal representation of products that can be scanned. This novel artificial neural network design would convert all inputs into a uniform ``category,’’ enabling the model to efficiently learn from a consistent data representation. After the training process, a conversion mechanism would transform test samples into this specific format, ensuring compatibility with the model's input requirements. An additional converter at the output stage would then revert the model's results back to their original form, whether it be visual, textual, or numerical. This strategy facilitates the creation of a versatile, self-adapting model capable of dynamically learning and adjusting based on the characteristics of incoming data.

\section{Conclusion}
\label{sec:conclusion}
In this article, we explored various learning algorithms and their significance across a wide range of applications. The core concepts of Artificial Intelligence (AI), Machine Learning (ML), Deep Learning (DL), hybrid models, Explainable AI (XAI), and their real-world applications were reviewed. We discussed some important subsets of machine learning algorithms that can be applied to solve real-world problems involving classification, prediction, and segmentation. We also explored the architecture of Convolutional Neural Networks (CNNs) and how they can be integrated with ML algorithms to develop hybrid models. The paper further addressed the vulnerability of learning algorithms to adversarial noise and its impact on model performance. Additionally, we investigated the integration of learning algorithms with Large Language Models (LLMs) for generating precise and contextual responses across various domains. Finally, we discussed the potential for future advancements, including the development of adaptive and dynamic networks capable of handling a broad range of intelligent tasks. In summary, this article provides a brief overview of learning algorithms, emphasizing their diverse applications, capabilities, and future directions.

\bibliographystyle{ieeetr}


\end{document}